\documentclass{article}

\usepackage{microtype}
\usepackage{graphicx}
\usepackage{subfigure}
\usepackage{booktabs} %

\usepackage{hyperref}

\usepackage[accepted]{icml2024}

\usepackage{amsmath}
\usepackage{amssymb}
\usepackage{mathtools}
\usepackage{amsthm}
\usepackage{amssymb, bm}
\usepackage{multirow}

\usepackage{xcolor}

\usepackage[capitalize,noabbrev]{cleveref}

\theoremstyle{plain}
\newtheorem{theorem}{Theorem}[section]
\newtheorem{proposition}[theorem]{Proposition}

\theoremstyle{definition}

\theoremstyle{remark}
\newtheorem{remark}[theorem]{Remark}

\usepackage[textsize=tiny]{todonotes}

\icmltitlerunning{Self-Attention through Kernel-Eigen Pair Sparse Variational Gaussian Processes}

\begin{document}

\twocolumn[
\icmltitle{Self-Attention through Kernel-Eigen Pair Sparse Variational Gaussian Processes}

\icmlsetsymbol{equal}{*}

\begin{icmlauthorlist}
\icmlauthor{Yingyi Chen}{equal,esat}
\icmlauthor{Qinghua Tao}{equal,esat}
\icmlauthor{Francesco Tonin}{epfl}
\icmlauthor{Johan A.K. Suykens}{esat}
\end{icmlauthorlist}

\icmlaffiliation{esat}{ESAT-STADIUS, KU Leuven, Belgium}
\icmlaffiliation{epfl}{LIONS, EPFL, Switzerland (most of the work was done at ESAT-STADIUS, KU Leuven)}

\icmlcorrespondingauthor{Yingyi Chen}{yingyi.chen@esat.kuleuven.be, chenyingyi076@gmail.com}

\icmlkeywords{Machine Learning, ICML}

\vskip 0.3in
]

\printAffiliationsAndNotice{\icmlEqualContribution} %

\begin{abstract}
While the great capability of Transformers significantly boosts prediction accuracy, it could also yield overconfident predictions and require calibrated uncertainty estimation, which can be commonly tackled by Gaussian processes (GPs).
Existing works apply GPs with symmetric kernels under variational inference to the attention kernel; however, omitting the fact that attention kernels are in essence asymmetric.
Moreover, the complexity of deriving the GP posteriors remains high for large-scale data.
In this work, we propose Kernel-Eigen Pair Sparse Variational Gaussian Processes (KEP-SVGP) for building uncertainty-aware self-attention where the asymmetry of attention kernels is tackled by Kernel SVD (KSVD) and a reduced complexity is acquired.
Through KEP-SVGP, 
\textit{i)} the SVGP pair induced by the two sets of singular vectors from KSVD w.r.t. the attention kernel fully characterizes the asymmetry;
\textit{ii)} using only a small set of adjoint eigenfunctions from KSVD, the derivation of SVGP posteriors can be based on the inversion of a diagonal matrix containing singular values, contributing to a reduction in time complexity;
\textit{iii)} an evidence lower bound is derived so that variational parameters and network weights can be optimized with it.
Experiments verify our excellent performances and efficiency on in-distribution, distribution-shift and out-of-distribution benchmarks.
\end{abstract}

\section{Introduction}
In recent years, Transformers \cite{vaswani2017attention} stand out among deep learning models, achieving state-of-the-art performances and excelling in feature learning in diverse applications \cite{brown2020language,dosovitskiy2021an,touvron2021training,wu2022flowformer}.
However, the large architecture capacities of Transformers could also lead to overconfident predictions \cite{guo2017calibration,mukhoti2020calibrating} with risks of robustness-related issues in safety-critical applications \cite{moon2020confidence,zhu2023openmix} where reliable uncertainty quantification can help.
Bayesian approaches allowing rich probabilistic interpretations of model predictions have been well studied on modern neural networks \cite{blundell2015weight,gal2016dropout,kendall2017uncertainties,salimbeni2017doubly,geifman2019bias,Zhang2020Cyclical}, where posterior inferences are often conducted in weight spaces \cite{foong2020expressiveness,ritter2021sparse,Coker2022WideMeanFieldBayesian}.
In Transformers, uncertainty estimation with 
Variational Inference (VI) \cite{graves2011practical} is relatively less studied.
Existing works include the studies using VI on layer weights \cite{tran2019bayesian,xue2021bayesian}, attention matrix \cite{fan2020bayesian,cinquin2022pathologies} and attention outputs \cite{liu2020simple,chen2023calibrating}, vital in providing reliable predictions.

Gaussian processes (GPs) \cite{rasmussen2006gaussian} serve as principal tools for uncertainty estimation within Bayesian inference.
Though GPs provide posterior distributions in closed forms, they are intractable for large datasets, e.g., long-sequence data for Transformers, as time complexity to the posterior GPs scale as $\mathcal{O}(N^3)$ where $N$ is the number of training samples.
Sparse Variational Gaussian Process (SVGP) \cite{titsias2009variational} deploying VI is proposed as an efficient alternative to classical GP.
It conducts posterior approximation based on a small set of $s$ ``inducing points (variables)'' yielding a reduction of time complexity from $\mathcal{O}(N^3)$ to $\mathcal{O}(Ns^2)$.
Recently, SVGPs are utilized \cite{chen2023calibrating} on attention outputs for uncertainty estimation.
However, we underscore that the self-attention kernel is in essence asymmetric \cite{tsai2019transformer,wright2021transformers,chen2023primal}, whereas SVGPs can only be characterized with symmetric kernels, resulting in a nontrivial gap in capturing the intrinsic rationale.

\begin{figure}[ht]
    \begin{center}
    \centerline{\includegraphics[width=\columnwidth]{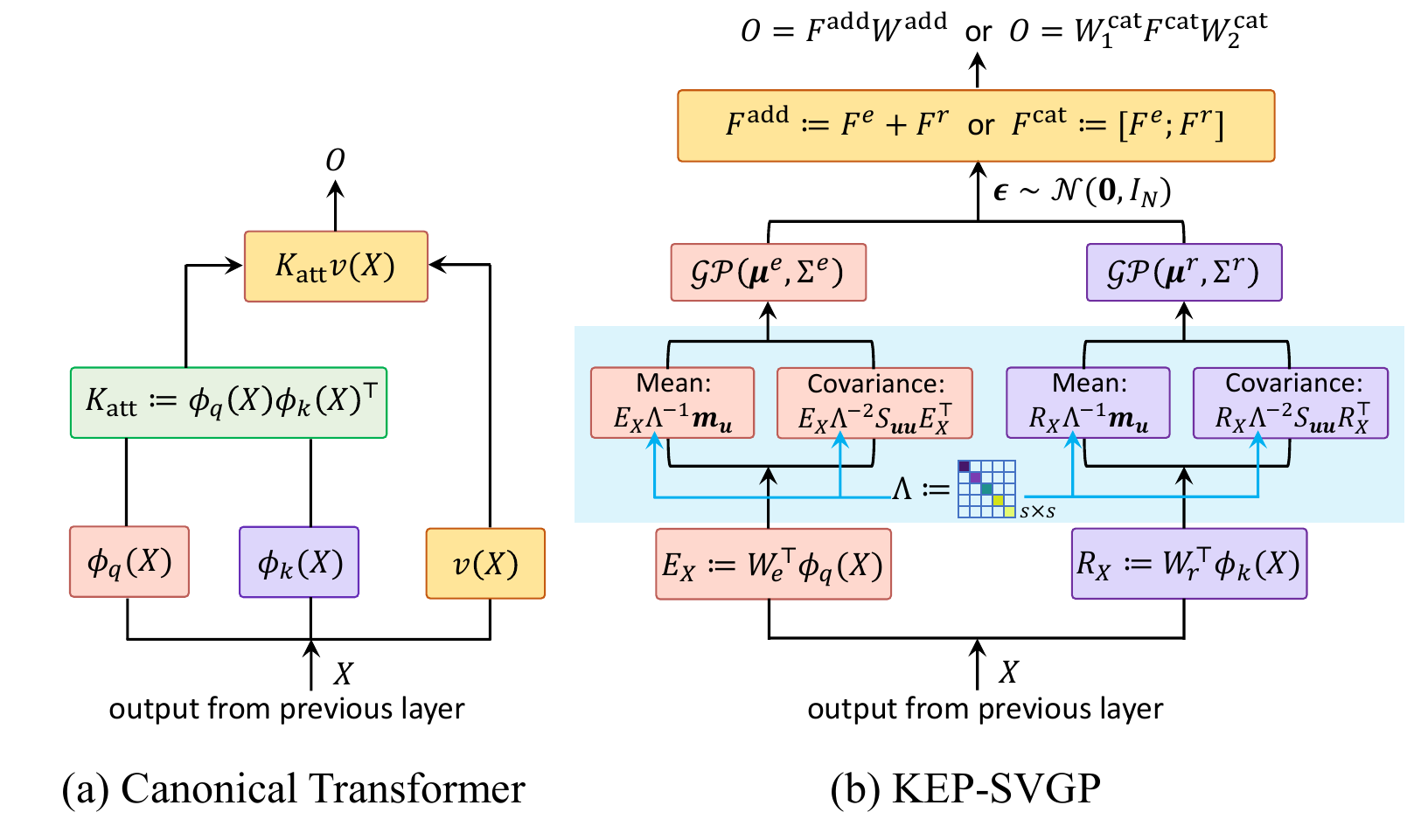}}
    \caption{Illustration of canonical self-attention and our KEP-SVGP in one layer.
    (a) The attention kernel $K_{\rm att}$ in canonical self-attention is induced by two different feature maps $\phi_q,\phi_k$ related to queries and keys; hence $K_{\rm att}$ is in essence asymmetric.
    (b) KEP-SVGP consists of one SVGP pair induced by the two sets of projection outputs based on $\phi_q,\phi_k$ from KSVD w.r.t. $K_{\rm att}$, which fully characterizes the asymmetry of self-attention in the posterior. 
    The posteriors are now approximated based on the inversion of a diagonal matrix $\Lambda$ containing top $s$ singular values,
    thereby of time complexity $\mathcal{O}(s)$.}
    \label{fig::workflow}
    \end{center}
    \vskip -0.3in
\end{figure}

\citet{chen2023primal} casts the asymmetric self-attention kernel in the framework of Kernel Singular Value Decomposition (KSVD) \cite{suykens2016svd,tao2023nonlinear}, which fully characterizes the asymmetry of the attention kernel through two sets of projection outputs w.r.t. both right and left singular vectors and can be efficiently optimized through an auxiliary loss.
In this paper, we propose Kernel-Eigen Pair Sparse Variational Gaussian Processes (KEP-SVGP) for building uncertainty-aware self-attention where the asymmetry of attention kernel is tackled by KSVD and a reduced time complexity in computing the SVGPs posterior is also acquired.
Specifically, through KEP-SVGP:
\vspace{-3mm}
\begin{itemize}
    \item \textit{Our SVGP pair induced by the left and right singular vectors of KSVD w.r.t.~the attention kernel matrix fully characterizes the asymmetry in the posterior.}
    This SVGP pair is obtained by setting the pair of adjoint eigenfunctions \cite{schmidt1907theorie,stewart1993early} w.r.t.~the asymmetric attention kernel to formulate ``inducing variables'', which is a technique named kernel-eigen features for SVGPs \cite{lazaro2009inter,leibfried2020tutorial}.
    Note that this technique has not yet been explored for large-architecture models, such as Transformers.
    \item \textit{We reduce the time complexity of the matrix inversion in SVGPs posterior approximation from $\mathcal{O}(s^3)$ to  $\mathcal{O}(s)$.}
    By using the singular vectors-induced SVGP pair, the posterior is now approximated based on the inversion of a truncated singular value matrix, e.g., corresponding to top-$s$ singular values, which is a diagonal matrix.
    \item \textit{An evidence lower bound (ELBO) tailored for KEP-SVGP is derived,} so that variational parameters and the network weights can be jointly optimized with ELBO.
    \item \textit{KEP-SVGP's efficacy and efficiency are experimentally verified} on in-distribution, distribution-shift and out-of-distribution benchmarks without sacrificing accuracy.\footnote{Code is at \href{https://github.com/yingyichen-cyy/KEP-SVGP}{https://github.com/yingyichen-cyy/KEP-SVGP}.}
\end{itemize}

\section{Related Work}
\paragraph{Uncertainty Estimation} 
Uncertainty estimation aims at providing confidence scores to the predictions, hence helping 
with both interpretability and trustworthiness of the models, and benefits downstream tasks, including failure prediction \cite{hendrycks2016baseline,li2024sure}, model calibration \cite{guo2017calibration}, OOD detection \cite{zhu2023openmix}, etc.
MSP \cite{hendrycks2016baseline} takes the maximum softmax probabilities from the softmax label distribution in deep neural network (DNN) classifiers as confidence scores.
Temperature Scaling \cite{guo2017calibration} is a post-hoc model calibration method which optimizes a single temperature parameter within the softmax in the DNN classifier on a validation set to calibrate predictions. 
Monte-Carlo Dropout (MC Dropout) \cite{gal2016dropout} casts dropout training in DNNs as approximate Bayesian inference in deep Gaussian processes, hence modelling uncertainty with dropout NNs.
Kronecker-factored last layer Laplace approximation (KFLLLA) \cite{kristiadi2020being} calibrates uncertainty on a ReLU network with asymptotic confidence of a last-layer Gaussian-approximated binary ReLU classifier with Laplace approximation.
Deep Ensembles \cite{lakshminarayanan2017simple} ensembles DNNs for well-calibrated uncertainty estimates.
Note that all the above mentioned methods can be easily implemented in different network architectures, such as Transformers.

\paragraph{Deep Gaussian Processes} 
\citet{damianou2013deep} introduces a deep hierarchy of GPs, where each layer is a GP or GP latent variable model. 
Rather than combining with DNNs, \citet{damianou2013deep} utilizes deep GP hierarchies and are limited to small datasets. 
\citet{aitchison2021deep} also focuses on the deep hierarchy of GPs by encompassing \citet{damianou2013deep} into deep kernel processes, where each GP layer with isotropic covariance kernel is connected to Wishart distribution and thus is fully characterized by the Gram matrix.
\citet{milsom2023convolutional} proposes convolutional deep kernel machine (C-DKM). Despite the naming, it belongs to deep GPs operating on Gram matrices \cite{aitchison2021deep} but focuses on convolutional kernels. 
Another category is deep kernel learning (DKL).
\citet{wilson2016stochastic} proposes stochastic variational deep kernel learning (SV-DKL), which introduces an extra GP layer to DNN backbones, where the DKL is in the sense of applying kernel machines on top of a DNN, which is agnostic in architectures. 
SV-DKL requires two-step training, i.e., the backbone pretraining and the finetuning with the extra GP layer. 
\citet{ober2021promises} further empirically investigates \citet{wilson2016stochastic}.
Discussions on KEP-SVGP's connections to deep GPs are provided in Section \ref{sec:svgp:eigen:features}.

\paragraph{Bayesian Transformers} 
\citet{tran2019bayesian} and \citet{xue2021bayesian} perform  variational inference (VI) on Transformers' layer weights, which can underfit the data, verified both empirically \cite{chen2023calibrating} and theoretically \cite{foong2020expressiveness,Coker2022WideMeanFieldBayesian}.
\citet{fan2020bayesian} and \citet{cinquin2022pathologies} perform VI on attention matrix where their experimental settings can be too restrictive for complex problems \cite{chen2023calibrating}.
VI can be also performed on attention outputs \cite{liu2020simple,chen2023calibrating}.
\citet{liu2020simple} fits an extra GP layer over the last layer output, which can be treated as a deep kernel learning in Transformer.
\citet{chen2023calibrating} proposes Sparse Gaussian Process Attention (SGPA) by formulating the self-attention with one SVGP.
Our KEP-SVGP is different from SGPA in 
\textit{i)} we use two SVGPs with two kernel-eigen features, hence taking the asymmetry in self-attention into consideration, while SGPA considers one vanilla SVGP;
\textit{ii)} we manage to reduce time complexity for self-attention to linear with the number of input data $N$, while SGPA remains $N^2$. More discussions on the comparisons of time complexity between KEP-SVGP and SGPA are provided in Section \ref{sec:svgp:eigen:features}.

\section{Background}

\subsection{Sparse Variational Gaussian Processes}
\label{subsec::SVGP}

\paragraph{Gaussian Processes}
\label{subsec::gps}
A GP \cite{rasmussen2006gaussian} represents a distribution, denoted by $\mathcal{GP}$, over real-valued functions $f(\cdot):\mathcal{X}\to\mathbb{R}$ defined on an input domain $\mathcal{X}\subset\mathbb{R}^d$.
A GP prior is characterized through two real-valued functions: a mean function $\mu(\cdot):\mathcal{X}\to\mathbb{R}$ which is often set to zero without loss of generality, and a symmetric positive-definite covariance function parameterized by a kernel function $\kappa(\cdot,\cdot):\mathcal{X}\times\mathcal{X}\to\mathbb{R}$.
When evaluating a GP at any finite number of inputs
$X=[\bm{x}_1,\ldots,\bm{x}_N]^\top$, $\bm{x}_i\in \mathcal{X}$, 
we obtain a Gaussian marginal distribution of function values $\bm{f}:=[f(\bm{x}_1),\ldots, f(\bm{x}_N)]^\top\in\mathbb{R}^N$, that is, 
\begin{equation}
    \text{Prior:}\,f(\cdot)\sim\mathcal{GP}(0,\kappa(\cdot,\cdot))
    \quad
    \Rightarrow
    \quad
    \bm{f}\sim\mathcal{N}(\bm{0}, K_{XX}),
\end{equation}
with $K_{XX}:=[\kappa(\bm{x}_i,\bm{x}_j)]\in\mathbb{R}^{N\times N}$. 
The training data is $(X,\bm{y}):=\{(\bm x_i, y_i)\}_{i=1}^N$ with $\bm{y}=[y_1,\ldots,y_N]^\top$, $y_i\in\mathbb{R}$
being the given outputs to the inputs $X$. 
With the likelihood $\bm{y}|\bm{f}\sim\mathcal{N}(\bm{f},\sigma^2I_N)$ being Gaussian, the posterior is also a GP.
Given the test inputs $X^*$, the posterior predictive distribution of $\bm{f}^*$ is
\begin{equation}
\label{eq::posterior_vanilla}
    \begin{split}
    & \mathrm{p}(\bm{f}^*|X^*,X,\bm{y})
    =\mathcal{N}\left(K_{X^*X}(K_{XX}+\sigma^2I_N)^{-1}\bm{y}\right., 
    \\
    & \left. K_{X^*X^*}-K_{X^*X}(K_{XX}+\sigma^2I_N)^{-1}K_{XX^*}\right).
    \end{split}
\end{equation}
However, \eqref{eq::posterior_vanilla} is intractable for large-scale data as the inversion of an $N\times N$ matrix is of time complexity $\mathcal{O}(N^3)$.

\paragraph{Sparse Variational Gaussian Processes}
SVGPs \cite{titsias2009variational} variationally approximate GP posteriors with a small set of $s$ supports, i.e., $(Z, \bm{u}):=\{(\bm z_m, u_m)\}_{m=1}^s$, $\bm z_m\in \mathcal{X}$, $u_m=f(\bm{z}_m)\in\mathbb{R}$ where the ``inducing variables'' $\bm{u}$ are evaluated at the ``inducing points'' $Z$.
In SVGPs, the mean $\bm{\mu}_{\bm{u}}$ is set to zero without loss of generality and the covariance matrix is $K_{ZZ}:=[\kappa(\bm{z}_i,\bm{z}_j)]\in\mathbb{R}^{s\times s}$. 
Other than considering the marginal distribution $\mathrm{p}(\bm{u})=\mathcal{N}(\bm{0},K_{ZZ})$, SVGPs give a variational distribution $\mathrm{q}(\bm{u})=\mathcal{N}(\bm{m}_{\bm{u}}, S_{\bm{u}\bm{u}})$, $\bm{m}_{\bm{u}}\in\mathbb{R}^{s}$, $S_{\bm{uu}}\in\mathbb{R}^{s\times s}$ \cite{leibfried2020tutorial}.
Thus, a marginal distribution over $\bm f$ can be obtained by $\mathrm{q}(\bm{f})=\int \mathrm{p}(\bm{f}|\bm{u})\mathrm{q}(\bm{u})\,\mathrm{d}\bm{u}$, which corresponds to the posterior whose distribution is also Gaussian:
\begin{equation}
\label{eq::approx_posterior}
    \begin{split}
    &\mathrm{q}(\bm{f})= \mathcal{N}\left(K_{XZ}K_{ZZ}^{-1}\bm{m}_{\bm{u}},\right.
    \\
    & \left. K_{XX}-K_{XZ}K_{ZZ}^{-1}(K_{ZZ}-S_{\bm{u}\bm{u}})K_{ZZ}^{-1}K_{ZX}\right),
   \end{split}
\end{equation}
where the kernel function values are $K_{XZ}:=[\kappa(\bm{x}_i,\bm{z}_j)]\in\mathbb{R}^{N\times s}$, 
$K_{ZX}:=K_{XZ}^\top$.
In inference, the approximate posterior distribution evaluated at test inputs can then be obtained with \eqref{eq::approx_posterior}. 
The optimization of SVGPs proceeds to maximize
the evidence lower bound (ELBO) $\mathbb{E}_{\mathrm{q}(\bm{f})}\left[\log\mathrm{p}\left(\bm{y}|\bm{f}\right)\right]- \text{KL}\left(\mathrm{q}(\bm{u})\|\mathrm{p}(\bm{u})\right)$ for the variational parameters $\bm{m}_{\bm{u}}$ and $S_{\bm{uu}}$ in the variational distribution $\mathrm{q}(\bm{u})$.
Detailed derivations are given in Appendix \ref{sec:appendix:svgp:vanilla}.

\paragraph{Kernel-Eigen Features for SVGPs}
Further in SVGPs, the ``inducing variables'' $\bm{u}$ can be alternatively chosen as a linear functional on $f(\cdot)$, which is called the ``inter-domain GPs'' \cite{lazaro2009inter,leibfried2020tutorial}, such that
\begin{equation}\label{eq:inducing:features}
u_m = \int f(\bm{x})\phi_m(\bm{x})\,\mathrm{d}\bm{x}, \quad m = 1, \ldots,s,
\end{equation}
where $\{\phi_{m}(\cdot)\}_{m=1}^s$ are the ``inducing features'' through the real-valued function $\phi_{m}(\cdot):\mathcal{X}\to\mathbb{R}$.
In particular, let $\phi_m(\cdot):=\nu_m(\cdot)$ be chosen as the $m$-th eigenfunction of the symmetric kernel  $\kappa(\cdot,\cdot)$ with eigenvalue $\lambda_m$, that is,
\begin{equation}
\label{eq:evd:integral}
    \lambda_m \nu_m(\cdot) = \int \kappa(\cdot,\bm{x})\nu_m(\bm{x})\,\mathrm{d}\bm{x}, \quad m = 1, \ldots, s.
\end{equation}
When evaluating the SVGP prior over a finite set $X\subset\mathcal{X}$ and its inducing points $Z$, the chosen eigenfunction $\nu_m(\cdot)$ for the ``inducing features'' $\phi_m(\cdot)$ in \eqref{eq:inducing:features} leads to an eigenvalue problem, which corresponds to the finite case of the integral equations w.r.t. the symmetric kernel function $\kappa(\cdot,\cdot)$ in \eqref{eq:evd:integral} \cite{NIPS2000_19de10ad}:
\begin{equation}
\label{eq:evd:svgp}
    K_{XX}H=H\Lambda,
\end{equation}
where $H:=[\bm{\nu}_1,\ldots,\bm{\nu}_s]\in\mathbb{R}^{N\times s}$ contains the eigenvectors to the top-$s$ nonzero eigenvalues of the kernel matrix $K_{XX}$, i.e., $\Lambda=\text{diag}\{\lambda_1,\ldots,\lambda_s\}$.
With $\mathrm{q}(\bm{u})=\mathcal{N}(\bm{m}_{\bm{u}},S_{\bm{uu}})$, the posterior distribution \eqref{eq::approx_posterior} in SVGPs with kernel-eigen features is then yielded as: 
\begin{equation}
\label{eq::approx_pos_sparse}
\begin{array}{c}
    \text{Prior:}
    \begin{pmatrix}
        \bm{f} \\ \bm{u}
    \end{pmatrix}
    \sim \mathcal{GP}
    \left( \bm{0},
    \begin{bmatrix}
        K_{XX} & H\Lambda
        \\
        \Lambda H^\top & \Lambda
    \end{bmatrix}
    \right)
    \vspace{0.15cm}
    \\
    \Downarrow \\
    \begin{split}
    & \mathrm{q}(\bm{f})
    =\mathcal{N}\left( 
    (H\Lambda)\Lambda^{-1}\bm{m}_{\bm{u}},\right.
    \\
    & \left. K_{XX} - (H\Lambda)\Lambda^{-1}(\Lambda-S_{\bm{u}\bm{u}})\Lambda^{-1}(\Lambda H^\top)
    \right).
    \end{split}
\end{array}
\end{equation} 

\begin{remark}\label{rmk:approx}
    By the compact SVD $K_{XX}=\sum_{i=1}^R \lambda_i \bm{\nu}_i \bm{\nu}_i^\top$ with $R$ the rank of $K_{XX}$,
    the covariance of the posterior in \eqref{eq::approx_pos_sparse} can be written as 
    $K_{XX} - H\Lambda H^\top + H S_{\bm{uu}}H^\top
    = U \Lambda_U U^\top + H S_{\bm{uu}}H^\top$,
    where $\Lambda_U=\text{diag}\{\lambda_{s+1},\ldots,\lambda_R\}$ are the smallest non-zero $(R-s)$ eigenvalues, and columns of $U$ are the corresponding eigenvectors.
    By the Eckart-Young theorem \cite{eckart1936}, $H\Lambda H^\top$ is the best rank-$s$ approximation to $K_{XX}$.
    For low-rank matrices, such as the self-attention \cite{wang2020linformer}, $\|U \Lambda_U U^\top\|^2_F$ is small, motivating to a faster-to-compute approximate posterior by
    $\tilde{\rm q}(\bm f)\sim\mathcal{N}\left( H\bm{m}_{\bm u}, H S_{\bm{uu}}H^\top\right)$.
    Validity of this approximation is numerically verified in Appendix \ref{appdx::low_rank}.
\end{remark}

Recall that regular SVGPs \eqref{eq::approx_posterior} give the posterior involving the inversion on $K_{ZZ}\in\mathbb{R}^{s\times s}$, and thereby  have a time complexity of $\mathcal{O}(s^3)$.
In contrast, with kernel-eigen features in SVGPs, the empirical covariance matrix w.r.t. $\bm{u}$ becomes diagonal, i.e., $\Lambda$, hence the time complexity of the matrix inversion is $\mathcal{O}(s)$, leading to a greater improvement in efficiency.
Detailed derivations are in Appendix \ref{sec:appendix:svgp:kernel_eigen}.

\subsection{Self-Attention as Asymmetric Kernel Machine} 
\label{eq:sec:2:attention}

\paragraph{Self-Attention corresponds to Asymmetric Kernel}
Let the input data sequence be $\{\bm{x}_i\}_{i=1}^N$, $\bm{x}_i\in\mathcal{X}$, self-attention formulates the queries $q(\bm{x}_i)=W_q\bm{x}_i$, $W_q\in\mathbb{R}^{d_q\times d}$, keys $k(\bm{x}_i)=W_k\bm{x}_i$, $W_k\in\mathbb{R}^{d_k\times d}$, and values $v(\bm{x}_i)=W_v\bm{x}_i$, $W_v\in\mathbb{R}^{d_v\times d}$, commonly with $d_q=d_k$. 
As pointed out in \citet{tsai2019transformer}, the 
attention matrix can be interpreted as a kernel matrix with entries depicting the asymmetric similarities between queries and keys:
\begin{equation}
\label{eq:kernel:attention}
    \kappa_{\rm att}(\bm{x}_i,\bm{x}_j):=\text{softmax}(\left<W_q\bm{x}_i,W_k\bm{x}_j\right>/\sqrt{d_k}),
\end{equation}
where $\kappa_{\rm att}(\cdot,\cdot):\mathcal{X}\times\mathcal{X}\to\mathbb{R}$ is the kernel yielding attention matrix
$K_{\rm att}:=[\kappa_{\rm att}(\bm{x}_i,\bm{x}_j)]\in\mathbb{R}^{N\times N}$. 
As $W_q\neq W_k$ generally,  we have $\left<W_q\bm{x}_i,W_k\bm{x}_j\right>\neq\left<W_q\bm{x}_j,W_k\bm{x}_i\right>$, so that the attention matrix is essentially asymmetric with $K_{ij}\neq K_{ji}$.
The canonical self-attention output in each head is denoted as $O:=[\bm{o}_1,\ldots,\bm{o}_N]^\top\in\mathbb{R}^{N\times d_v}$ with
\begin{align} 
\label{eq::output_cano}
    \bm{o}_i=\sum\nolimits_{j=1}^N v(\bm{x}_j)\kappa_{\rm att}(\bm{x}_i,\bm{x}_j),\quad 
    i=1,\ldots,N.
\end{align}
Kernel-based approaches have become popular in studying the attention \cite{choromanski2021rethinking,nguyen2022fourierformer,chi2022kerple,nguyen2023a}.  
However, they resort to the techniques with symmetric kernels where the inputs are $\{q(\bm{x}_i)\}_{i=1}^N$, $\{k(\bm{x}_j)\}_{i=1}^N$.
Differently, the following \citet{chen2023primal} works directly on the asymmetric kernel function and is grounded on the original input $\{\bm{x}_i\}_{i=1}^N$.

\paragraph{Self-Attention with Kernel SVD}
\label{subsubsec::ksvd}
\citet{chen2023primal} formulates the self-attention mechanism with KSVD \cite{suykens2016svd,tao2023nonlinear} which allows asymmetric kernels, and derives a primal-dual framework  to represent the attention outputs and the optimization. 
The asymmetric kernel for self-attention is introduced as $\kappa_{\rm att}(\bm x_i, \bm x_j)=\left<\phi_q(\bm x_i),\phi_k(\bm x_j)\right>$ with $\phi_q(\cdot):\mathcal{X}\to\mathbb{R}^p$ and $\phi_k(\cdot):\mathcal{X}\to\mathbb{R}^p$ related to queries and keys, respectively.
The primal-dual representations of self-attention with KSVD give:
\begin{equation}
\label{eq::primal_dual}
 \begin{array}{l}
    \begin{array}{rll}
        \text{Primal:} & 
        \left\{
        \begin{array}{rl}
            e(\bm{x})=W_{e}^\top\phi_q(\bm{x}), 
            \\
            r(\bm{x})=W_{r}^\top\phi_k(\bm{x}),
        \end{array}
        \right.
        \vspace{1mm}
        \\
        \text{Dual:} &
        \left\{
        \begin{array}{l}
            {e}(\bm{x})=\sum\nolimits_{j=1}^N\bm h_{r_j}\kappa_{\rm att}(\bm{x},\bm{x}_j),
            \\
            {r}(\bm{x})=\sum\nolimits_{i=1}^N\bm h_{e_i}\kappa_{\rm att}(\bm{x}_i,\bm{x}),
        \end{array}
        \right.
    \end{array}
\end{array}
\end{equation}
where $e(\bm{x}), r(\bm{x})\in\mathbb{R}^s$ are the projections related to queries and keys, whose variances are maximized under KSVD as shown in the objective \eqref{eq::ksvd_loss}. 
Primal variables $W_e, W_r\in\mathbb{R}^{p\times s}$ serve as the projection weights,
and dual variables 
$H_e:=[\bm h_{e_1}, \ldots, \bm h_{e_N}]^\top$, 
$H_r:=[\bm h_{r_1}, \ldots, \bm h_{r_N}]^\top\in\mathbb R^{N\times s}$ 
are column-wisely the left and right singular vectors of the attention matrix $K_{\rm att}$.
Note that the canonical self-attention outputs in \eqref{eq::output_cano} corresponds to the dual representation of the projection score $e(\bm x)$ in \eqref{eq::primal_dual} once setting $\bm{h}_{r_j}:=v(\bm{x}_j)$.

To fully exploit the asymmetry in self-attention kernel matrix, \citet{chen2023primal} proposes Primal-Attention, which concatenates both projections $e(\bm x)$, $r(\bm x)$ w.r.t. right and left singular vectors, such that $F_i:=[e(\bm{x}_i);r(\bm{x}_i)]=[W_{e}^\top\phi_q(\bm{x}_i); W_{r}^\top\phi_k(\bm{x}_i)]$. 
With the KKT conditions, the stationary solutions to KSVD yield
a zero-value objective, as proved in Lemma 4.2 in \citet{chen2023primal}. 
Thus, the KSVD optimization in Primal-Attention can be flexibly implemented by minimizing an auxiliary regularization loss:
\begin{equation}
\label{eq::ksvd_loss}
    \begin{array}{rl}
    & \min\limits_{W_e,W_r,\Lambda}\mathcal{L}_{\rm KSVD}
    := \big[-\frac{1}{2}\sum\nolimits_{i=1}^Ne(\bm{x}_i)^\top\Lambda^{-1}e(\bm{x}_i) 
    \\
    & - \frac{1}{2}\sum\nolimits_{j=1}^Nr(\bm{x}_j)^\top\Lambda^{-1}r(\bm{x}_j)
    \vspace{0.1cm}
    +\text{Tr}(W_e^\top W_r)\big]^2,
    \end{array}
\end{equation}
seeking for the projections with maximal variances w.r.t. the two sets of singular vectors, where $\Lambda\in\mathbb{R}^{s\times s}$ is a positive diagonal matrix of the {top-$s$ singular values}. 
Primal-Attention avoids computing the attention kernel matrix in the dual by deploying the primal representations with greater efficiency. 
Details of \citet{chen2023primal} are in Appendix \ref{sec::background::appendix}.

\section{KEP-SVGP for Self-Attention}
In this section, the method of KEP-SVGP is illustrated in Section \ref{sec:svgp:eigen:features}, where two branches of SVGPs with the adjoint kernel-eigen feature pair are considered based on KSVD to capture asymmetry in self-attention.
Section \ref{sec:optimization} provides the optimization of KEP-SVGP.

\subsection{Kernel-Eigen Pair SVGP}
\label{sec:svgp:eigen:features}
In (SV)GPs, the kernel function is required to be symmetric, whereas the attention in Transformers is in essence asymmetric.
As shown in \citet{chen2023primal}, the asymmetric kernel in self-attention can be fully characterized by two sets of projections under the KSVD framework. 
To variationally model the outputs with the asymmetric attention kernel, we use the pair of adjoint eigenfunctions in the integral equations w.r.t. the attention kernel for the kernel-eigen features, so as to formulate the ``inducing variables'' in SVGPs.

\paragraph{Pair of Adjoint Eigenfunctions for Self-Attention}
With asymmetric $\kappa_{\rm att}(\cdot, \cdot)$ \eqref{eq:kernel:attention}, its adjoint eigenfunctions $\nu_{e_m}(\cdot)$, $\nu_{r_m}(\cdot)$ regarding the eigenvalue $\lambda_m$ \cite{schmidt1907theorie} satisfy
\begin{equation}
\label{eq:integral:svd}
    \begin{array}{l}
    \lambda_m \nu_{e_m}(\cdot) = \int\kappa_{\rm att}(\cdot, \bm z) \nu_{r_m} (\bm z) \, \mathrm{d}\bm z,
    \\
    \lambda_m \nu_{r_m}(\cdot) = \int  \kappa_{\rm att}(\bm x, \cdot)\nu_{e_m}(\bm x)\, \mathrm{d}\bm x.
    \end{array}
\end{equation}
The finite-sample cases to the integrals in \eqref{eq:integral:svd} correspond to the compact SVD on an asymmetric attention kernel matrix $K_{\rm att}\in\mathbb{R}^{N\times N}$, i.e., KSVD \cite{tao2023nonlinear}, 
where the approximation to $\nu_{e_m}(\cdot)$, $\nu_{r_m}(\cdot)$ leads to the left and right singular vectors.
To differentiate from the symmetric cases, eigenvalue $\lambda_m$ in \eqref{eq:integral:svd} with asymmetric kernels are named as singular values \cite{stewart1993early}.
In self-attention, this yields the shifted eigenvalue problem \cite{lanczos1958linear,suykens2016svd} w.r.t.~the attention matrix $K_{\rm att}$ \cite{chen2023primal}:
\begin{equation}
\label{eq::shifted_eigen}
    K_{\rm att} H_r=H_e\Lambda, 
    \quad
    K_{\rm att}^\top H_e=H_r\Lambda,
\end{equation} 
where $\Lambda=\text{diag}\{\lambda_1,\ldots,\lambda_s\}$ contains the top-$s$ nonzero singular values of $K_{\rm att}$, $H_e, H_r\in\mathbb{R}^{N\times s}$ are kernel-eigen features defined in Section \ref{subsubsec::ksvd}.
However, the kernel matrix in \eqref{eq::shifted_eigen} is asymmetric, hence inconsistent with the symmetry requirements of SVGPs in \eqref{eq:evd:svgp}.

Two sets of integral equations w.r.t.~a pair of symmetric kernels can be introduced to equivalently characterize the integral equations in \eqref{eq:integral:svd} \cite{schmidt1907theorie,stewart1993early}:
\begin{equation}
\label{eq:integral:svd:symmetric}
    \begin{array}{l}
      \lambda^2_m  \nu_{e_m}(\cdot) = \int
     \kappa_e(\cdot,\bm{z})
      \nu_{e_m} (\bm z) \, \mathrm{d}\bm{z},  
      \\
    \lambda^2_m \nu_{r_m}(\cdot) = 
    \int  
    \kappa_r(\bm{x},\cdot)
    \nu_{r_m}(\bm x) \, \mathrm{d}\bm x,
    \end{array}
\end{equation}
where $\kappa_e(\cdot,\bm{z}):=\int \kappa_{\rm att}(\cdot, \bm y)\kappa_{\rm att}(\bm{z}, \bm y) \,\mathrm{d}\bm y$,
$\kappa_r(\bm{x},\cdot):=\int \kappa_{\rm att}(\bm y, \bm{x})\kappa_{\rm att}(\bm y, \cdot)\,\mathrm{d}\bm y$ correspond to two symmetric and positive definite kernels and $\lambda_m^2$ is the eigenvalue in the induced two sets of integral equations. 
Thus, we can deploy a SVGP pair with kernel-eigen features \eqref{eq:evd:integral} to variationally model the self-attention outputs.

Given the training data sequences in self-attention, the finite-sample cases to integrals \eqref{eq:integral:svd:symmetric} give  two eigendecompositions w.r.t. symmetric kernels $\kappa_e$, $\kappa_r$: 
\begin{equation}
\label{eq::two_symm_eigen}
    (K_{\rm att}K_{\rm att}^\top) H_e=H_e\Lambda^2, 
    \,  
    (K_{\rm att}^\top K_{\rm att}) H_r=H_r\Lambda^2.
\end{equation} 
The asymmetric attention matrix $K_{\rm att}$ can be fully characterized by the symmetric $K_{\rm att}K_{\rm att}^\top$ and $K_{\rm att}^\top K_{\rm att}$ with kernel-eigen feature pair $H_e$, $H_r$  serving as the ``inducing features'' in the resulting two SVGPs.
Detailed derivations of \eqref{eq::two_symm_eigen} are given in Appendix \ref{sec:two:evd:appendix}.

\paragraph{SVGPs with Kernel-Eigen Features Pair}
In basic setups of (SV)GPs, the processes are of single-output as in Section \ref{subsec::gps} with $f(\cdot):\mathcal{X}\to\mathbb{R}$.
For the multi-dimensional attention outputs, we consider independent 
multi-output Gaussian processes \cite{leibfried2020tutorial} where a separate SVGP is specified for each output dimension \cite{salimbeni2017doubly,chen2023calibrating}.  
Following the KSVD framework with Primal-Attention setups \cite{chen2023primal}, we model the $s$-dimensional attention outputs with the two sets of projections for capturing the asymmetry, denoted as $F^e_{[d]}:=F^e[:,d]$, $F^r_{[d]}:=F^r[:,d]\in \mathbb{R}^N$,  $d=1,\ldots,s$, w.r.t. $e(\bm x)$, $r(\bm x)$ in \eqref{eq::primal_dual}, respectively. 
Therefore, with \eqref{eq::two_symm_eigen}, a SVGP pair w.r.t. the two symmetric kernels $K_{\rm att}K_{\rm att}^\top$ and $K_{\rm att}^\top K_{\rm att}$ induced by the asymmetric $K_{\rm att}$ is established. With the kernel-eigen feature pair $H_e$, $H_r$ on $\bm{f}_{[d]}^e,\bm{f}_{[d]}^r\in\mathbb{R}^N$, 
similar as \eqref{eq::approx_pos_sparse} with Remark \ref{rmk:approx} 
our SVGPs are attained as follows:
\begin{equation*}
    \begin{array}{c}
    \text{Prior:}
    \begin{pmatrix}
        \bm{f}^e_{[d]} \\ \bm{u}^e_{[d]}
    \end{pmatrix}
    \sim \mathcal{GP}
    \left( \bm{0},
    \begin{bmatrix}
        K_{\rm att}K_{\rm att}^\top & H_e\Lambda^2
        \\
        \Lambda^2H_e^\top & \Lambda^2
    \end{bmatrix}
    \right)
    \\
    \Downarrow
    \\
    \tilde{\mathrm{q}}(\bm{f}^e_{[d]})=\mathcal{N}
    \Big(
    \underbrace{E_X\Lambda^{-1}\bm{m}_{\bm{u},[d]}}_{\bm{\mu}^e:=\bm{m}^e_{[d]}},
    \,
    \underbrace{E_X\Lambda^{-2}S_{\bm{uu},[d]} E_X^\top}_{\Sigma^e:=L^e_{[d]}{L^e_{[d]}}^\top}
    \Big)
    \end{array}
\end{equation*}
\begin{equation}
\label{eq::e:r:score_sgp}
    \begin{array}{c} 
    \text{Prior:}
    \begin{pmatrix}
        \bm{f}^r_{[d]} 
        \\ \bm{u}^r_{[d]}
    \end{pmatrix}
    \sim \mathcal{GP}
    \left( \bm{0},
    \begin{bmatrix}
        K_{\rm att}^\top K_{\rm att} & H_r\Lambda^2
        \\
        \Lambda^2H_r^\top & \Lambda^2
    \end{bmatrix}
    \right)
    \\
    \Downarrow
    \\
    \tilde{\mathrm{q}}(\bm{f}^r_{[d]})=\mathcal{N}
    \Big(
    \underbrace{R_X\Lambda^{-1}\bm{m}_{\bm{u},[d]}}_{\bm{\mu}^r:=\bm{m}^r_{[d]}},
    \,
    \underbrace{R_X\Lambda^{-2}S_{\bm{uu},[d]} R_X^\top}_{\Sigma^r:=L^r_{[d]}{L^r_{[d]}}^\top}
    \Big)
    \end{array}
\end{equation}
with variational distributions on $\bm{u}^e_{[d]},\bm{u}^r_{[d]}\in\mathbb{R}^s$:
\begin{equation}
\label{eq::same_var_dis}
    \bm{u}^e_{[d]}, \bm{u}^r_{[d]}\sim\mathcal{N}(\bm{m}_{\bm{u},{[d]}}, S_{\bm{uu},{[d]}}),
\end{equation}
where $E_X:=[e(\bm{x}_i),\ldots,e(\bm{x}_N)]^\top\in\mathbb{R}^{N\times s}$ and
$R_X:=[r(\bm{x}_i),\ldots,r(\bm{x}_N)]^\top \in \mathbb R^{N\times s} $
are the projection matrices w.r.t. right and left singular vectors of KSVD in \eqref{eq::primal_dual}, and 
$\bm{m}_{\bm{u}} \in\mathbb{R}^{s\times s}$, 
$S_{\bm{uu}} \in\mathbb{R}^{s\times s \times s}$ are the variational parameters with
$\bm{m}_{\bm{u},{[d]}}:=\bm{m}_{\bm{u}}[:,d]\in\mathbb{R}^s$, 
$S_{\bm{uu},[d]}:=S_{\bm{uu}}[:,:,d] \in\mathbb{R}^{s\times s}$
corresponding to the $d$-th output dimension.
Note that we set $\bm{u}^e_{[d]}$,$\bm{u}^r_{[d]}$ 
from the same variational distribution since the priors in SVGP pair in \eqref{eq::e:r:score_sgp} share the same marginal distribution of the ``inducing variables''.
Detailed derivations of our SVGP pair \eqref{eq::e:r:score_sgp} are given in Appendix \ref{sec:appendix:twin:svgp}.

Based on the approximate posteriors in \eqref{eq::e:r:score_sgp}, the outputs of the two SVGPs are obtained by the reparameterization trick \cite{salimbeni2017doubly}:
\begin{equation}
\label{eq:two:outputs}
    F_{[d]}^e= \bm{m}^e_{[d]} +L^e_{[d]}\bm{\epsilon}, 
    \quad
    F_{[d]}^r=\bm{m}^r_{[d]}+L^r_{[d]}\bm{\epsilon},
\end{equation} 
with $\bm{\epsilon}\sim\mathcal{N}(0,I_N)$, where $\bm{m}^e_{[d]}$, $\bm{m}^r_{[d]}$ are the means in \eqref{eq::e:r:score_sgp} 
and
$L^e_{[d]}:=E_X\Lambda^{-1}L_{\bm{uu},[d]}$,
$L^r_{[d]}:=R_X\Lambda^{-1}L_{\bm{uu},[d]}$ are Cholesky factors of the approximate posterior covariances $\Sigma^e,\Sigma^r$ in \eqref{eq::e:r:score_sgp}, and $L_{\bm{uu},[d]}$ is the Cholesky factor of $S_{\bm{uu},[d]}$.

\paragraph{Merging the SVGPs}
Utilizing the SVGP pair in \eqref{eq::e:r:score_sgp} to preserve the asymmetric $K_{\rm att}$, we propose two schemes to merge two SVGPs outputs in \eqref{eq:two:outputs}:
\begin{equation}
\label{eq:merge:output}
    \begin{array}{rl}
        \text{Addition:} &  F_{[d]}:=F_{[d]}^e + F_{[d]}^r \in\mathbb{R}^{N},
         \vspace{0.1cm}
         \\
         \text{Concatenation:} & F_{[d]}:=[F_{[d]}^e;F_{[d]}^r ]\in\mathbb{R}^{2N}.
    \end{array}
\end{equation}
To align with the $d_v$ dimensions in  standard Transformer architectures \eqref{eq::output_cano}, our $s$-dimensional outputs are applied with linear projections, similar to \citet{chen2023primal}. 
Specifically, the final outputs $O \in \mathbb R^{N \times d_v}$ of the attention layer are:
$O:= F^{\text{add}}W^{\text{add}}$ for the addition, $O:= W^{\text{cat}}_1 F^{\text{cat}} W^{\text{cat}}_2$ for the concatenation,
where $F=[F_{[1]},\ldots,F_{[s]}]$ is the output matrix of our merged SVGPs with projection matrices  
$W^{\text{add}}\in\mathbb{R}^{s\times d_v}$,
$W^{\text{cat}}_1\in\mathbb{R}^{N\times 2N}$ and $W^{\text{cat}}_2\in\mathbb{R}^{s\times d_v}$.
Both schemes can be applied to data with fixed sequence lengths common in computer vision tasks, while for those with varying sequence lengths common in language modelling, we turn to the addition scheme. 
Note that for the concatenation scheme, we can replace $W_1^{\text{cat}}$ with $AB^\top$, $A\in\mathbb{R}^{N\times s}$, $B\in\mathbb{R}^{2N\times s}$ to maintain the linear complexity with $N$. More details can be found in Appendix \ref{sec:ab:appendix}.

\paragraph{Discussions on Time Efficiency}
\label{subsec::efficiency}
The time complexity of MSP \cite{hendrycks2016baseline}, i.e., the canonical Transformer with softmax self-attention, is $\mathcal{O}(BN^2)$, where $B$ is the batch size, $N$ is the data sequence length.
Recently, \citet{chen2023calibrating} employs SVGPs for self-attention where Bayesian inference is performed in the attention output space to calibrate uncertainty.
\citet{chen2023calibrating} proposes:
\textit{i)} standard SGPA, which is based on {regular} SVGPs, with a time complexity of $\mathcal{O}(BN^3)$;
\textit{ii)} decoupled SGPA, which sets $s$ global ``inducing points'', with time complexity $\mathcal{O}(BN^2s + s^3)$.
However, decoupled SGPA still scales quadratically w.r.t. the sequence length and needs the matrix inversion with $\mathcal{O}(s^3)$.
In contrast, our posterior distribution in \eqref{eq::e:r:score_sgp} involves the matrix multiplication with  $\mathcal{O}(BNs^2)$ and the inversion of the $s\times s$ diagonal matrix $\Lambda$  with only $\mathcal{O}(s)$, leading to $\mathcal{O}(BNs^2+Bs)$.
To alignment with the hidden dimensions in standard Transformers, after \eqref{eq:merge:output}, our concatenation merging scheme takes the matrix multiplication of $\mathcal{O}(N^2s)$.
In practice, we commonly have $s<N$ with $s$ being distinctively smaller than $N$, so that, omitting the effect of batch size, \textit{the main time complexities in MSP, SGPA, and KEP-SVGP scale as $\mathcal{O}(N^2)$, $\mathcal{O}(N^2s)$, and $\mathcal{O}(Ns^2)$ for addition scheme, $\mathcal{O}(N^2s)$ for concatenation scheme, respectively.}
In practice, we pertain considerable efficiency advantages for both our merging schemes, as we by default apply KEP-SVGP to the last layer, yielding better performances,  instead of all layers as SGPA.
Experiments on training time efficiency are in Table \ref{tab::efficiency} and Appendix \ref{sec:time_eff:appendix}.

\begin{table*}[t]
    \caption{Mean and standard deviations on CIFAR-10, CIFAR-100, IMDB, CoLA benchmarks.
    Experimental results are reported over five trials, with the best mean results shown in bold.
    ACC, AUROC, FPR95, ECE and Brier are percentages, 
    AURC is $\times 10^3$, NLL is $\times 10$.
    }
    \label{tab::in_dist}
    \begin{center}
    \resizebox{\textwidth}{!}{
    \begin{tabular}{ccccccccc}
        \toprule
        Dataset & Method & ACC/MCC $\uparrow$ & AURC $\downarrow$ & AUROC $\uparrow$ & FPR95 $\downarrow$ & ECE $\downarrow$ & NLL $\downarrow$ & Brier $\downarrow$ 
        \\ \midrule
        \multirow{7}{*}{
        \begin{tabular}[c]{@{}c@{}}CIFAR-10\\ \cite{krizhevsky2009learning}\end{tabular}}  
        & MSP \cite{hendrycks2016baseline}       
        & 83.50$\pm$0.43 & 42.60$\pm$1.84 & 86.15$\pm$0.35 & 66.51$\pm$2.19 & 12.87$\pm$0.29 & 11.13$\pm$0.40 & 28.62$\pm$0.74
        \\
        & Temperature Scaling \cite{guo2017calibration}              
        & 83.50$\pm$0.43 & 40.47$\pm$1.63 & 86.55$\pm$0.36 & 65.10$\pm$2.23 & 9.50$\pm$0.25 & 6.70$\pm$0.20 & 26.05$\pm$0.68
        \\
        & MC Dropout \cite{gal2016dropout}   
        & 83.69$\pm$0.51 & 41.36$\pm$1.45 & 86.18$\pm$0.28 & 66.49$\pm$1.96 & 12.48$\pm$0.43 & 10.35$\pm$0.41 & 28.09$\pm$0.73
        \\ 
        & KFLLLA \cite{kristiadi2020being}
        & 83.54$\pm$0.45 & 40.12$\pm$1.65 & 86.70$\pm$0.50 & 63.13$\pm$1.75 & {\bf 1.51}$\pm$0.18 & {\bf 5.08}$\pm$0.10 & 23.75 0.57
        \\
        &SV-DKL \cite{wilson2016stochastic}
        & 83.82$\pm$0.58 & 39.78$\pm$1.91 & 86.57$\pm$0.38 & 65.02$\pm$1.33 & 11.32$\pm$0.55 & 7.88$\pm$0.57 & 27.03$\pm$0.96
        \\
        & KEP-SVGP (ours)      
        & 84.70$\pm$0.61 & 35.15$\pm$2.65 & 87.20$\pm$0.65 & 64.93$\pm$1.41 & 10.60$\pm$0.45 & 8.00$\pm$0.56 & 25.45$\pm$1.05  
        \\ \cmidrule(lr){2-9} 
        & Deep Ensembles \cite{lakshminarayanan2017simple}    
        & 86.43 & 27.76 & 88.64 & 60.72 & 9.98 & 7.40 & 22.89       
        \\
        & KEP-SVGP Ensembles (ours)       
        & \bf 87.62	& \bf 22.56	& \bf 89.64 & \bf 56.70 & 8.19 & 5.61 &	\bf 20.08  
        \\ \midrule
        \multirow{7}{*}{\begin{tabular}[c]{@{}c@{}}CIFAR-100\\ \cite{krizhevsky2009learning}\end{tabular}} 
        & MSP \cite{hendrycks2016baseline}             
        & 52.82$\pm$0.53 & 229.25$\pm$4.41 & 82.01$\pm$1.93 & 75.45$\pm$0.83 & 30.97$\pm$0.61 & 33.21$\pm$1.13 & 74.89$\pm$1.03           
        \\
        & Temperature Scaling \cite{guo2017calibration} 
        & 52.82$\pm$0.53 & 223.83$\pm$4.18 & 82.47$\pm$0.34 & 71.79$\pm$0.97 & 17.36$\pm$0.68 & 21.82$\pm$0.54 & 64.92$\pm$0.84
        \\
        & MC Dropout \cite{gal2016dropout}      
        & 53.37$\pm$0.62 & 224.01$\pm$4.82 & 81.44$\pm$0.17 & 75.12$\pm$1.12 & 30.09$\pm$0.80 & 31.78$\pm$1.22 & 73.51$\pm$1.24 
        \\
        & KFLLLA \cite{kristiadi2020being}
        & 51.35$\pm$0.64 & 263.59$\pm$4.65 & 79.62$\pm$0.19 & 71.86$\pm$1.30 & 37.98$\pm$4.61 & 27.21$\pm$1.66 & 82.84$\pm$3.82
        \\
        & SV-DKL \cite{wilson2016stochastic}
        & 53.00$\pm$0.71 & 228.10$\pm$8.35 & 81.41$\pm$0.63 & 72.22$\pm$1.63 & {\bf 1.59}$\pm$0.22 & {\bf 17.28}$\pm$0.31 & 60.07$\pm$0.87
        \\
        & KEP-SVGP (ours)          
        & 55.02$\pm$0.83 & 209.75$\pm$6.20 & 81.71$\pm$0.30 & 74.03$\pm$0.90 & 27.80$\pm$0.57 & 28.38$\pm$0.45 & 69.91$\pm$1.12 
        \\ \cmidrule(lr){2-9} 
        & Deep Ensembles \cite{lakshminarayanan2017simple}     
        & 59.65 & 165.27 & 83.84 & 71.40 & 24.41 & 22.93 & 62.01       
        \\ 
        & KEP-SVGP Ensembles (ours)         
        & \bf 62.45 & \bf 144.63 & \bf 84.56 & \bf 70.68 & 21.01 & 19.64 & \bf 56.63   
        \\ \midrule
        \multirow{7}{*}{\begin{tabular}[c]{@{}c@{}}IMDB\\ \cite{maas2011learning}\end{tabular}} 
        & MSP \cite{hendrycks2016baseline}             
        & 88.17$\pm$0.52 & 35.27$\pm$3.04 & 82.29$\pm$0.87 & 71.41$\pm$1.57 & 4.01$\pm$1.36 & 3.10$\pm$0.26 & 17.88$\pm$0.95         
        \\
        & Temperature Scaling \cite{guo2017calibration}              
        & 88.17$\pm$0.52 & 35.27$\pm$3.04 & 82.29$\pm$0.87 & 71.08$\pm$1.55 & {\bf 1.05}$\pm$0.70 & 2.89$\pm$0.12 & 17.40$\pm$0.80 
        \\
        & MC Dropout \cite{gal2016dropout}      
        & 88.34$\pm$0.65 & 34.62$\pm$3.17 & 82.24$\pm$0.83 & 71.65$\pm$2.03 & 2.66$\pm$1.84 & 2.97$\pm$0.27 & 17.47$\pm$1.19  
        \\
        & KFLLLA \cite{kristiadi2020being}
        & 88.17$\pm$0.52 & 35.20$\pm$3.01 & 82.31$\pm$0.86 & 71.07$\pm$1.51 & 19.13$\pm$0.73 & 4.38$\pm$0.07 & 25.88$\pm$0.63 
        \\
        & SV-DKL \cite{wilson2016stochastic}
        & 88.86$\pm$1.04 & 59.84$\pm$18.90 & 73.20$\pm$5.56 & 69.91$\pm$3.68 & 7.31$\pm$8.27 & 3.38$\pm$0.82 & 19.62$\pm$5.18
        \\
        & SGPA \cite{chen2023calibrating}            
        & 88.36$\pm$0.75 & 33.14$\pm$3.46 & 82.78$\pm$0.44 & 70.85$\pm$2.46 & 5.52$\pm$0.46 & 3.40$\pm$0.10 & 18.05$\pm$0.81          
        \\
        & KEP-SVGP (ours)          
        & 89.01$\pm$0.14 & 30.69$\pm$0.69 & 83.22$\pm$0.31 & 68.15$\pm$0.95 & 3.72$\pm$0.81 & 3.00$\pm$0.13 & 16.56$\pm$0.25  
        \\ \cmidrule(lr){2-9} 
        & Deep Ensembles \cite{lakshminarayanan2017simple}     
        & 89.57 & 28.69 & 83.45 & 67.69 & 2.42 & \bf 2.68 & \bf 15.60  
        \\ 
        & KEP-SVGP Ensembles (ours)         
        & \bf 89.68 & \bf 27.79 & \bf 83.56 & \bf 67.54 & 3.43 & 2.84 & 15.68       
        \\ \midrule
        \multirow{7}{*}{\begin{tabular}[c]{@{}c@{}}CoLA\\ \cite{warstadt2019neural}\end{tabular}} 
        & MSP \cite{hendrycks2016baseline}             
        & 26.93$\pm$1.38 & 205.47$\pm$7.62 & 64.55$\pm$0.86 & 89.86$\pm$1.29 & 23.84$\pm$2.23 & 14.45$\pm$2.83 & 52.15$\pm$2.43 
        \\
        & Temperature Scaling \cite{guo2017calibration}              
        & 26.93$\pm$1.38 & 205.46$\pm$7.61 & 64.55$\pm$0.91 & 90.09$\pm$0.77 & 18.98$\pm$3.33 & 8.72$\pm$1.18 & 47.59$\pm$2.85  
        \\
        & MC Dropout \cite{gal2016dropout}      
        & 26.41$\pm$1.87 & 203.93$\pm$8.34 & 65.15$\pm$0.76 & 88.58$\pm$0.53 & 23.33$\pm$2.16 & 13.74$\pm$2.64 & 51.35$\pm$2.43
        \\
        & KFLLLA \cite{kristiadi2020being}
        & 26.90$\pm$1.31 & 204.31$\pm$8.57 & 64.60$\pm$0.96 & 90.06$\pm$0.74 & {\bf 2.51}$\pm$1.09 & {\bf 5.94}$\pm$0.04 & {\bf 40.52}$\pm$0.38
        \\
        & SV-DKL \cite{wilson2016stochastic}
        & 26.65$\pm$1.38 & 235.76$\pm$10.03 & 62.14$\pm$1.36 & 89.94$\pm$1.79 & 18.13$\pm$8.92 & 8.07$\pm$1.86 & 48.45$\pm$6.41
        \\
        & SGPA \cite{chen2023calibrating}            
        & 26.15$\pm$1.12 & 210.03$\pm$6.30 & 64.18$\pm$0.68 & 90.35$\pm$1.47 & 16.48$\pm$0.79 & 8.76$\pm$0.34 & 45.77$\pm$0.54 
        \\
        & KEP-SVGP (ours)          
        & 30.54$\pm$1.61 & 186.66$\pm$8.50 & 65.16$\pm$0.86 & 88.39$\pm$0.83 & 15.89$\pm$3.48 & 8.54$\pm$1.66 & 43.55$\pm$2.99 
        \\ \cmidrule(lr){2-9} 
        & Deep Ensembles \cite{lakshminarayanan2017simple}     
        & 27.35 & 184.96 & 67.02 & 87.93 & 22.82 & 12.45 & 49.45       
        \\ 
        & KEP-SVGP Ensembles (ours)         
        & \bf 31.02 & \bf 164.06 & \bf 67.88 & \bf 85.18 & 14.96 & 7.40 & 40.68 
        \\ \bottomrule
    \end{tabular}}
    \end{center}
    \vspace{-6mm}
\end{table*}

\paragraph{Discussions on Connections to Deep GPs}
KEP-SVGP proposes a variational self-attention mechanism performing Bayesian inference in the outputs of multi-head attention blocks. 
\textbf{Relevance:} When multiple self-attention layers in a Transformer are replaced by KEP-SVGP, the model can be categorized as a specialized deep GPs; 
when only the last self-attention layer is replaced by KEP-SVGP, the model can be categorized as a stochastic variational model upon deep kernel learning with Transformer backbones.
\textbf{Differences:}
\textit{i)} As attention matrix is asymmetric and cannot be simply treated as a GP covariance matrix, we propose two GP branches to capture the asymmetry. 
Current literature in deep GPs has not considered this asymmetric case observed in transformers. 
\textit{ii)} Rather than adding an extra GP layer on top of a DNN \cite{wilson2016stochastic,ober2021promises}, KEP-SVGP replaces the last self-attention layer without imposing extra layers for variational modelling. 
\textit{iii)} Compared to the two-step training in SV-DKL \cite{wilson2016stochastic}, KEP-SVGP trains from scratch. 
\textit{iv)} C-DKM \cite{milsom2023convolutional} is computationally demanding due to the large Gram matrices in each layer, while KEP-SVGP is efficient. 
Hence, some deep GPs methods are not comparable to KEP-SVGP in transformers due to different setups. 
Experiments on SV-DKL are in Section \ref{sec::Exp} with transformer for fair comparisons.

\subsection{Optimization of KEP-SVGP}
\label{sec:optimization}
In optimization, we derive the ELBO objective for training the variational parameters involved in our SVGP pair.
For Transformers with $N_{\text{h}}$ heads in $L$ attention layers applied with KEP-SVGP, we denote $\{F^{l}\in\mathbb{R}^{N\times (N_{\text{h}}d_v)}\}_{l=1}^L$ as the output of the $l$-th KEP-SVGP layer following the convention of linearly concatenating the heads. 
Since single-output SVGPs \cite{leibfried2020tutorial} are employed as explained in \eqref{eq:two:outputs}, we can perform the variational inference on each output dimension of the attention heads before concatenating them, rather than the inference directly on the multi-head attention. 
With $\{\bm{u}^{l,n_{\text{h}}}\}_{l=1,n_{\text{h}}=1}^{L,N_{\text{h}}}$ for the SVGPs in our KEP-SVGP, the ELBO is formulated as: 
\begin{equation}
\label{eq::elbo_spg}
    \begin{array}{ll}
    \max\limits_{\Theta,\{\bm{m}_{\bm{u}}, S_{\bm{uu}}\}}\mathcal{L}_{\text{ELBO}} 
     :=\mathbb{E}_{\mathrm{q}\left(F^L|F^0\right)} 
    \big[\log\mathrm{p}(Y|F^L)\big] 
    \\
    -\sum\limits_{l=1}^{L}\sum\limits_{{n_{\text{h}}=1}}^{{N_{\text{h}}}}
    \mathbb{E}_{\mathrm{q}(F^{l-1})}
    \big[
        \text{KL}\big(
        \mathrm{q}(\bm{u}^{l,{n_{\text{h}}}}|F^{l-1})
        \|
        \mathrm{p}(\bm{u}^{l,{n_{\text{h}}}}|F^{l-1}) 
        \big)
    \big]
    \end{array}
\end{equation} 
where $\Theta$ denotes all network weights including ${W_e, W_r}$ and $\Lambda$, $Y$ contains the labels of the input data $F^0:=X_{\text{in}}$.
In \eqref{eq::elbo_spg}, the first item in $\mathcal{L}_{\text{ELBO}}$ corresponds to the objective of the learning task, such as the cross-entropy loss, while the second term of the Kullback–Leibler divergence balances the distance between the prior and variational distribution of the inducing variables $\bm u^{l,{n_{\text{h}}}}$.
Since $\bm{u}^e$, $\bm{u}^r$ share the same marginal prior and variational distributions conditioned on each $F^{l-1}$, we can consider one $\bm{u}$ for the KL divergence term.
The KL divergence involved in the ELBO of our KEP-SVGP is depicted in Proposition \ref{prop:elbo}, with detailed derivations in Appendix \ref{sec:elbo:appendix}.

\begin{proposition}\label{prop:elbo}
The Kullback–Leibler divergence in the ELBO objective \eqref{eq::elbo_spg} {is equal to}
\begin{equation}\label{eq:kl:obj}
    \begin{array}{cl}
         & \frac{1}{2}\sum\limits_{d=1}^{s} \left[ \mathrm{Tr}(\Lambda^{-2}S_{\bm{uu},[d]}) 
         + \bm{m}_{\bm{u},[d]}^\top\Lambda^{-2}\bm{m}_{\bm{u},[d]} 
         \right.
         \\
         & \left.+\log\frac{|\Lambda^{2}|}{|S_{\bm{uu},[d]}|} - s\right]
    \end{array}
\end{equation}
where $\Lambda\in\mathbb{R}^{s\times s}$ is diagonal whose inversion is of $\mathcal{O}(s)$.
\end{proposition}

The training objective of KEP-SVGP is $\min \,-\mathcal{L}_{\text{ELBO}} + \eta \mathcal{L}_{\rm KSVD}$,
where $\eta>0$ is the regularization constant. 
In our objective, we also incorporate loss $\mathcal{L}_{\rm KSVD}$ in KSVD given in \eqref{eq::ksvd_loss}, ensuring that $H_e, H_r$ in \eqref{eq::two_symm_eigen} 
are kernel-eigen features defined in Section \ref{subsubsec::ksvd} for SVGPs.
Monte-Carlo sampling is used to compute $\mathcal{L}_{\text{ELBO}}$, where function values are generated iteratively by passing through each layer with the reparameterization trick \cite{kingma2013auto}.

\section{Experiments} \label{sec::Exp}
\paragraph{Datasets and Baselines}
We conduct empirical evaluations on benchmarks including
\textit{i)} computer vision: CIFAR-10, CIFAR-100 \cite{krizhevsky2009learning};
\textit{ii)} language modelling: IMDB sentiment analysis \cite{maas2011learning}, CoLA linguistic acceptability prediction \cite{warstadt2019neural}.
We compare our KEP-SVGP with 
\textit{i)} single-model methods: 
maximum softmax probability score (MSP) \cite{hendrycks2016baseline},
Temperature Scaling \cite{guo2017calibration},
Monte-Carlo Dropout (MC Dropout) \cite{gal2016dropout},
Kronecker-factored last layer Laplace approximation (KFLLLA) \cite{kristiadi2020being},
and SGPA \cite{chen2023calibrating};
\textit{ii)} ensemble method:
we compare our KEP-SVGP Ensembles with Deep Ensembles \cite{lakshminarayanan2017simple}.
In the experiments, we set the concatenation merging scheme for computer vision datasets, and the addition merging scheme for language modelling datasets.
Unless specified, we replace the last-layer self-attention with KEP-SVGP, as this simple setup already achieves better performances with improved efficiency.

\begin{table}[t]
    \caption{Performance under distribution shift. The averaged results for 15 kinds of corruption under five different 
    {levels of perturbation severity} are reported.}
    \vspace{-3mm}
    \label{tab::dis_shift}
    \begin{center}
    \resizebox{\columnwidth}{!}{
    \begin{tabular}{cccccccc}
        \toprule
        Method 
        & ACC $\uparrow$ 
        & AURC $\downarrow$ 
        & AUROC $\uparrow$& FPR95 $\downarrow$ & ECE $\downarrow$ & NLL $\downarrow$ & Brier $\downarrow$ 
        \\ \midrule
        \multicolumn{8}{c}{CIFAR-10-C} 
        \\ 
        MSP             
        & 69.17 & 151.07& 78.71 & 77.98 & 24.75 & 24.27 & 53.99 
        \\
        MC Dropout  
        & 69.26 & 150.11 & 78.70 & 78.01 & 24.26 & 22.79 & 53.39
        \\ 
        KFLLLA
        & 69.17 & 146.55 & 79.59 & 75.11 & \bf 6.94 & \bf 9.72 & \bf 43.17
        \\
        SV-DKL
        & 68.95 & 150.90 & 79.10 & 76.70 & 22.43 & 16.45 & 51.94
        \\
        KEP-SVGP (ours)      
        & 69.71 & 144.90 & 79.40 & 77.13 & 22.03 & 18.30 & 50.70 
        \\ \cmidrule(lr){2-8}
        Deep Ensembles  
        & \bf 73.93 & \bf 114.14 & 81.47 & 73.98 & 20.14 & 17.40 & 44.82
        \\
        KEP-SVGP Ensembles (ours)       
        & 73.67 & 115.65 & \bf 81.54 & \bf 73.85 & 18.34 & 14.17 & 43.42
        \\ \midrule
        \multicolumn{8}{c}{CIFAR-100-C} 
        \\ 
        MSP             
        & 39.19	& 394.14 & 76.64 & 79.52 & 40.53 & 48.98 & 96.31
        \\
        MC Dropout  
        & 39.62 & 389.91 & 76.67 & 79.27 & 39.73 & 47.29 & 95.12
        \\ 
        KFLLLA
        & 38.00 & 419.64 & 76.41 & 77.18 & 27.43 & 31.33 & 88.19 
        \\
        SV-DKL
        & 40.17 & 379.79 & 77.84 & 76.40 & \bf 19.48 & \bf 27.75 & \bf 78.76
        \\
        KEP-SVGP (ours)      
        & 39.69 & 391.65 & 76.57 & 78.93 & 37.82 & 43.83 & 93.13
        \\ \cmidrule(lr){2-8}
        Deep Ensembles  
        & \bf 46.33 & \bf 312.35 & \bf 78.90 & 76.73 & 33.40 & 35.91 & 82.98
        \\
        KEP-SVGP Ensembles (ours)       
        & 46.31 & 315.69 & 78.76 & \bf 76.38 & 31.21 & 33.03 & 80.93
        \\ \bottomrule
    \end{tabular}}
    \end{center}
    \vspace{-3mm}
\end{table}

\begin{figure}[t]
    \vskip -0.1in
    \begin{center}
    \centerline{\includegraphics[width=\columnwidth]{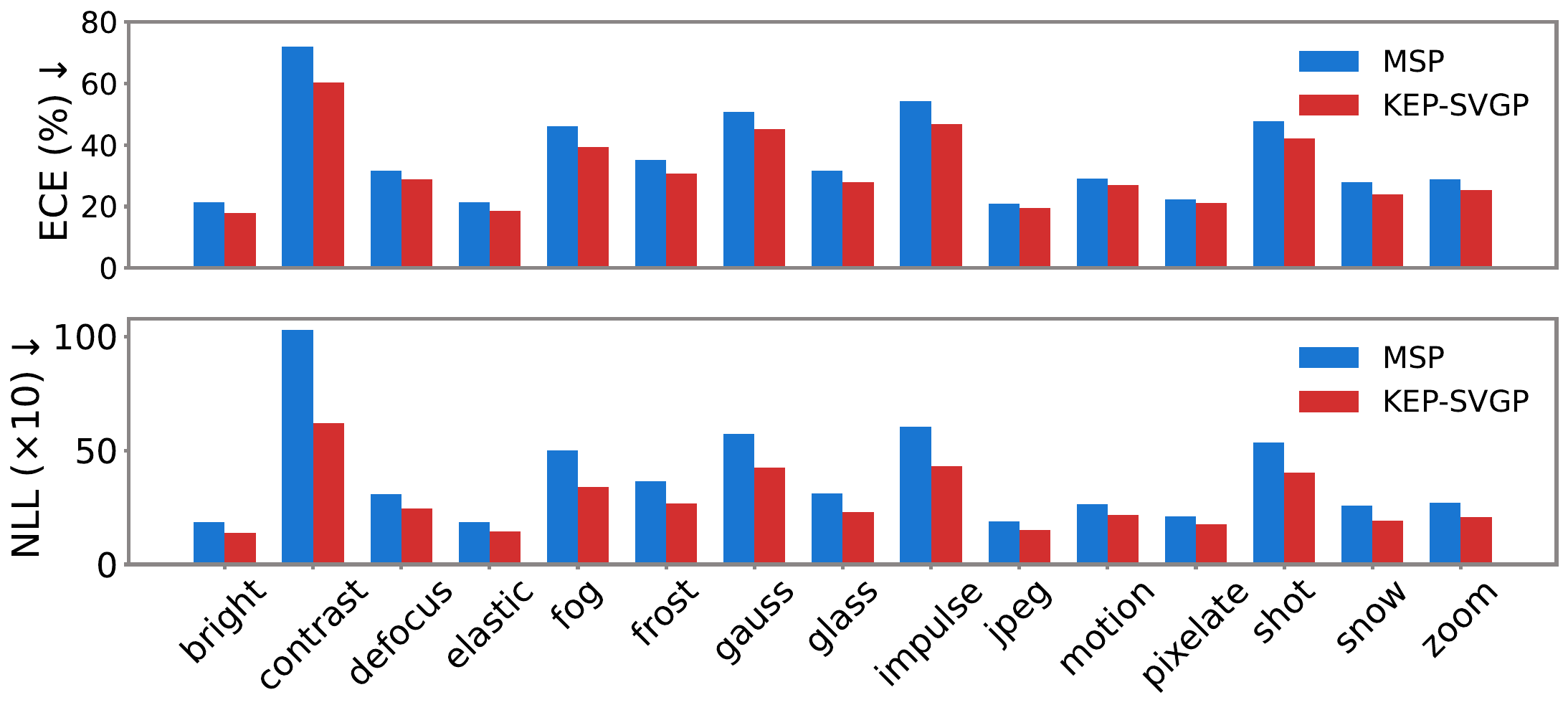}}
    \caption{Comparisons of our KEP-SVGP with MSP under distribution shift.
    Performance on 15 types of corruption under the severity level of 5 is reported, where models are trained on CIFAR-10/ViT and tested on CIFAR-10-C.}
    \label{fig::cifar10c}
    \end{center}
    \vskip -0.3in
    \vspace{-3mm}
\end{figure}

\paragraph{Training Configurations and Evaluation Metrics}
For both CIFAR-10, CIFAR-100, we train 7-layer Vision Transformer (ViT) \cite{dosovitskiy2021an}, optimized by Adam with batch size 128 and a cosine learning rate 
initialized with $10^{-3}$ for 300 epochs.
Following \citet{chen2023calibrating}, for IMDB, we adopt one-layer Transformer, trained with batch size 32 with a cosine learning rate initialized with $10^{-3}$ for 20 epochs;
for CoLA, we adopt a two-layer Transformer trained with batch size 32, an initial cosine learning rate of $5\times10^{-4}$ for 50 epochs. 
For evaluations, in addition to test accuracy (ACC), we consider a variety of metrics widely used for failure prediction and uncertainty calibration, presenting more comprehensive analyses:
\textit{i)} failure prediction:
the area under risk coverage curves (AURC) \cite{geifman2017selective}, 
the area under the receiver operating characteristic curve (AUROC) \cite{davis2006relationship},
FPR95 that returns FPR at 95\% TPR;
\textit{ii)} uncertainty calibration: 
expected calibration error (ECE) \cite{naeini2015obtaining},
negative predictive log-likelihood (NLL), 
and Brier score \cite{brier1950verification}.
More implementation details including choices of $s$, $\eta$, kernel functions, Monte-Carlo sampling for ELBO during inference are given in Appendix \ref{sec:setups:appendix}.

\begin{table}[t]
    \caption{OOD detection performance with AUROC (\%) and AUPR (\%). The average results over five trials are reported.}
    \label{tab::ood}
    \vspace{-3mm}
    \begin{center}
    \resizebox{\columnwidth}{!}{
    \begin{tabular}{ccccccc}
        \toprule
        ID & \multicolumn{3}{c}{CIFAR-10} & \multicolumn{3}{c}{CIFAR-100} 
        \\ \cmidrule(lr){2-4} \cmidrule(lr){5-7}
        OOD & SVHN & CIFAR-100 & LSUN & SVHN & CIFAR-10 & LSUN   
        \\ \midrule
        \multicolumn{7}{c}{AUROC $\uparrow$}      
        \\
        MSP 
        & 86.56	& 81.50 & 87.48
        & 75.83	& 67.14	& 74.97     
        \\
        MC Dropout 
        & 86.56 & 81.67 & 88.19
        & 76.62 & 67.54 & 74.94      
        \\
        KFLLLA 
        & 75.95 & 75.67 & 80.00
        & 72.81 & 65.37 & 71.25   
        \\
        SV-DKL
        & 75.48 & 76.81 & 82.02 
        & 74.35 & 65.72 & 72.03
        \\
        KEP-SVGP (ours)
        & 84.75 & 82.32	& 91.50
        & 79.98 & 67.51 & 78.22     
        \\  \cmidrule(lr){2-7}
        Deep Ensembles 
        & \bf 90.74	& 85.22 & 90.25  
        & 79.49	& 70.09 & 77.93        
        \\
        KEP-SVGP Ensembles (ours) 
        & 88.15 & \bf 85.36 & \bf 93.24  
        & \bf 84.16 & \bf 70.44 & \bf 81.28       
        \\ \midrule
        \multicolumn{7}{c}{AUPR $\uparrow$}      
        \\
        MSP 
        & 81.34	& 83.30 & 89.08
        & 65.85 & 68.42	& 78.74
        \\
        MC Dropout 
        & 81.89 & 83.50 & 89.69
        & 67.03 & 68.93 & 78.81
        \\
        KFLLLA 
        & 66.58 & 78.51 & 83.22 
        & 58.98 & 67.50 & 74.42    
        \\
        SV-DKL
        & 64.68 & 78.79 & 84.71	
        & 59.63 & 68.83 & 74.91
        \\
        KEP-SVGP (ours) 
        & 79.05	& 84.07	& 92.77
        & 71.57 & 68.83 & 81.65     
        \\ \cmidrule(lr){2-7}
        Deep Ensembles 
        & \bf 87.79 & 86.75 & 91.62
        & 71.38	& 71.30 & 82.05     
        \\
        KEP-SVGP Ensembles (ours) 
        & 84.35 & \bf 86.78 & \bf 94.29 
        & \bf 77.69 & \bf 71.68 & \bf 84.70       
        \\ \bottomrule
    \end{tabular}}
    \end{center}
\end{table}

\subsection{Comparison Results}
\begin{figure}[t]
    \vskip -0.1in
    \begin{center}
    \centerline{\includegraphics[width=\columnwidth]{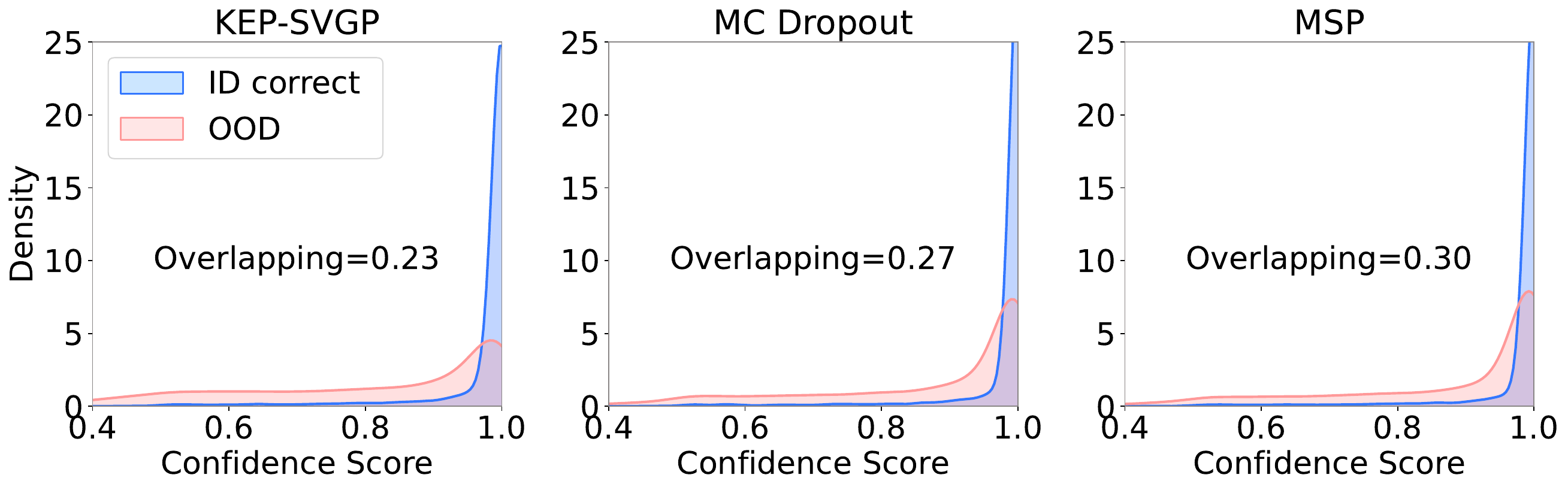}}
    \caption{KEP-SVGP leads to better confidence separation between ID correct and OOD samples.}
    \label{fig::ood_conf}
    \end{center}
    \vskip -0.3in
    \vspace{-2mm}
\end{figure}

\begin{table*}[t]
    \caption{Comparisons of performance and efficiency on a single NVIDIA Tesla V100 SXM2 32 GB. Results are reported over five trials.}
    \label{tab::efficiency}
    \begin{center}
    \resizebox{\textwidth}{!}{
    \begin{tabular}{cccccccccccccc}
        \toprule
        \multirow{2}{*}{Method} 
        & \multirow{2}{*}{\begin{tabular}[c]{@{}c@{}}Time\\Complexity\end{tabular}} 
        & \multicolumn{3}{c}{CIFAR-10}
        & \multicolumn{3}{c}{CIFAR-100}
        & \multicolumn{3}{c}{IMDB}
        & \multicolumn{3}{c}{CoLA}
        \\ \cmidrule(lr){3-5} \cmidrule(lr){6-8} \cmidrule(lr){9-11} \cmidrule(lr){12-14}
        & & ACC $\uparrow$ & NLL $\downarrow$ & s/Epoch 
        & ACC $\uparrow$ & NLL $\downarrow$ & s/Epoch 
        & ACC $\uparrow$ & NLL $\downarrow$ & s/Epoch 
        & MCC $\uparrow$ & NLL $\downarrow$ & s/Epoch 
        \\ \midrule
        MSP  
        & $\mathcal{O}(N^2)$
        & 78.11$\pm$0.10 & 13.40$\pm$0.07 & 29.58 
        & 52.16$\pm$0.50 & 43.90$\pm$0.42 & 29.76 
        & 88.17$\pm$0.52 & 3.10$\pm$0.26 & 16.65
        & 26.93$\pm$1.38 & 14.45$\pm$2.83 & 23.09         
        \\
        SGPA  
        & $\mathcal{O}(N^2s)$
        & 77.87$\pm$0.12 & 6.97$\pm$0.02 & 137.28
        & 53.02$\pm$0.36 & 25.64$\pm$0.41 & 288.04
        & 88.36$\pm$0.75 & 3.40$\pm$0.10 & 1662.36
        & 26.15$\pm$1.12 & 8.76$\pm$0.34 & 28.05           
        \\ \cmidrule(lr){2-14}
        KEP-SVGP (ours) 
        & {$\mathcal{O}(Ns^2)$}
        & {\bf 78.27}$\pm$0.30 & {\bf 6.29}$\pm$0.06 & 30.97
        & {\bf 56.26}$\pm$0.70 & {\bf 20.10}$\pm$1.10 & 32.21
        & {\bf 89.01}$\pm$0.14 & {\bf 3.00}$\pm$0.13 & 32.30
        & {\bf 30.54}$\pm$1.61 & {\bf 8.54}$\pm$1.66 & 23.95    
        \\ \bottomrule   
    \end{tabular}}
    \end{center}
    \vskip -0.2in
    \vspace{-2mm}
\end{table*}

\begin{table}[t]
    \caption{Ablation on {the asymmetry of
    our SVGP pair} on CoLA. Results are reported over five trials.}
    \label{tab::layer_branches_ablation}
    \vskip -0.2in
    \begin{center}
    \resizebox{\columnwidth}{!}{
    \begin{tabular}{cccccccc}
    \toprule
    Pairing & MCC $\uparrow$& AURC $\downarrow$& AUROC $\uparrow$& FPR95 $\downarrow$& ECE $\downarrow$& NLL $\downarrow$& Brier $\downarrow$
    \\ \midrule
    MSP
    & 26.93 & 205.47 & 64.55 & 89.86 & 23.84 & 14.45 & 52.15
    \\ \cmidrule(lr){2-8}
    $F_{[d]}^e + F_{[d]}^e$
    & 28.71 & 199.48 & 64.05 & 89.36 & 21.85 & 13.15 & 49.99  
    \\
    $F_{[d]}^r + F_{[d]}^r$
    & 28.10 & 196.65 & 64.66 & 90.01 & 18.99 & 10.76 & 47.68
    \\ \cmidrule(lr){2-8}
    $F_{[d]}^e + F_{[d]}^r$
    & \bf 29.31 & \bf 189.08 & \bf 64.88 & \bf 88.51 & \bf 16.08 & \bf 9.02 & \bf 44.16   
    \\ \bottomrule
    \end{tabular}}
    \end{center}
    \vskip -0.2in
    \vspace{-3mm}
\end{table}

\paragraph{Uncertainty Awareness on In-distribution Data}
In Table \ref{tab::in_dist}, we evaluate the in-distribution performances on four benchmarks where results on CoLA are measured with Matthew correlation coefficient (MCC) \cite{matthews1975comparison,warstadt2019neural}. 
Among single-model methods, KFLLLA and Temperature Scaling  are two post-hoc approaches designed for calibration and show good performances w.r.t. ECE, NLL and Brier.
However, the performance of KFLLLA is not stable across tasks, e.g., it has the highest ECE on CIFAR-100, while not able to improve much regarding the failure prediction metrics.
Our KEP-SVGP has consistently better performance across tasks.
Compared with MSP, our method not only improves upon ACC/MCC, but significantly reduces the AURC, NLL, and other comparing metrics.
Although SV-DKL using two-step training shows better performance in some metrics on CIFAR-100, KEP-SVGP has overall better performance regarding all benchmarks.
Notably, compared to the latest method for uncertainty calibration on Transformers (SGPA), KEP-SVGP distinctively surpasses its performances w.r.t. both failure prediction and calibration metrics, together with significantly improved efficiency as later shown in Table \ref{tab::efficiency}.
Deep Ensembles integrates five independently trained models and achieves more advantageous results than single-model methods. 
Compared to Deep Ensembles, KEP-SVGP Ensembles still outperforms it on all datasets and w.r.t. most metrics.
Moreover, KEP-SVGP can serve as a good complement for further improvement in calibration, given in Appendix \ref{sec:cal_boost:appendix}.

\paragraph{Robustness on Distribution-shift Data}
We consider CIFAR-10-C and CIFAR-100-C \cite{hendrycks2019benchmarking} in Table \ref{tab::dis_shift}, which are corrupted data of CIFAR-10/100 containing 15 types of input corruptions over 5 levels of corruption severity.
Models are trained on clean CIFAR-10/100 and evaluated on corrupted CIFAR-10/100-C.
As Temperature Scaling is designed for in-distribution uncertainty only, we omit its comparison herein.
Among almost all single-model methods, KEP-SVGP is with best ACC. 
Although KFLLLA has good calibration metrics on both datasets, its ACC is equal to or lower than that of MSP, indicating that its good calibration performance comes at the cost of ACC. 
SV-DKL is a strong method on CIFAR-100-C, however, its ACC on CIFAR-10-C is even lower than the MSP baseline, which is less desirable.
KEP-SVGP has good ACC with more stable uncertainty metrics.
Figure \ref{fig::cifar10c} shows the comparisons to MSP built upon canonical softmax-based Transformer where KEP-SVGP improves the model robustness under various corruptions.
With ensembles, Deep Ensembles is a classical method admitted with strong robustness against distribution shift. 
Notably, KEP-SVGP Ensembles provides better results than Deep Ensembles on more metrics, especially on CIFAR-10-C. 
As we are not aiming at achieving SOTAs in all metrics/setups, current results can verify KEP-SVGP's robustness under distribution shift.

\paragraph{Out-of-distribution Detection}
An effective estimator with reliable confidence is expected to well separate out-of-distribution (OOD) data and misclassified in-distribution (ID) data from correct predictions.
Hence, we consider OOD detection, where CIFAR-10/100 are taken as the ID data with evaluations on each other, SVHN \cite{netzer2011reading} and LSUN \cite{yu2015lsun}.
In Table \ref{tab::ood}, KEP-SVGP can overall boost the AUROC and AUPR over baselines.
Note that though KFLLLA has good performance on the tasks of in-distribution calibration and distribution-shift datasets, it cannot handle OOD detection well, while our method maintains its effectiveness.
Moreover, our KEP-SVGP Ensembles achieves the best performance almost in all cases, e.g., the boost on CIFAR-100$\to$SVHN is $4.67\%$ in AUROC, $6.31\%$ in AUPR over Deep Ensembles, which is quite substantial.
Figure \ref{fig::ood_conf} shows that KEP-SVGP leads to less confidence overlap between OOD and correct ID compared with MSP, MC Dropout, which is desirable. 

\subsection{Time Complexity}\label{subsec::time_efficiency}
Table \ref{tab::efficiency} gives the 
training time efficiency discussed 
in Section \ref{subsec::efficiency}.
We adopt same architectures in \citet{chen2023calibrating} on all datasets for fair comparisons with SGPA.
Compared to the latest SVGPs-based counterpart SGPA, KEP-SVGP significantly reduces the computational time over all benchmarks, which is consistent with the analytical results on time complexity.
Compared to MSP, it is reasonable that KEP-SVGP takes comparable or sometimes slightly longer training time as $\mathcal{O}(Ns^2)$ can be sometimes larger than $\mathcal{O}(N^2)$ with the chosen rank $s$.
KEP-SVGP distinctively outperforms MSP on all datasets and metrics, only with a slightly extra training time.
More results are in Appendix \ref{sec:time_eff:appendix}.

\subsection{Ablation Study}
In Table \ref{tab::layer_branches_ablation}, we investigate the effectiveness of leveraging our SVGPs for characterizing the asymmetric self-attention. 
We adopt a two-layer Transformer with all layers substituted by KEP-SVGP, which can also investigate our model applied to deep layers.
We consider addition scheme for merging the outputs of SVGPs, i.e., $F_{[d]}^e + F_{[d]}^r$, and compare with the symmetric cases, which only include a single SVGP under the same architecture, i.e., either $F_{[d]}^e + F_{[d]}^e$ or $F_{[d]}^r + F_{[d]}^r$. 
Results show that we achieve performance gains by leveraging our SVGP pair conceived from the asymmetry on the self-attention kernel.
Moreover, the 2-layer KEP-SVGP performs slightly inferior than 
transformers with only the last attention layer replaced by KEP-SVGP in Table \ref{tab::in_dist}.
This can be due to the fact that shallow layer may not necessarily enjoy a low-rank property \cite{chen2023primal}.
Nevertheless, all variants of
our 2-layer KEP-SVGP in Table \ref{tab::layer_branches_ablation}
clearly outperform the MSP baseline, further verifying the effectiveness of our SVGPs, especially the asymmetry in our KEP-SVGP.
More ablation results are provided in Appendix \ref{appdx::ablations}.

\section{Conclusion}
In this work, we propose a novel variational modelling for realizing more reliable self-attention outputs through two branches of SVGPs, which leverages the pair of adjoint eigenfunctions w.r.t. the asymmetric attention kernel to formulate a pair of kernel-eigen features for ``inducing features'' in SVGPs.
First, we fully characterize the intrinsic asymmetry of the attention kernel by utilizing the adjoint pair of SVGPs.
Second, by deploying KSVD into SVGPs, 
we manage to reduce the time complexity of posterior processes approximation significantly.
Third, we tailor the ELBO for optimizing the variational parameters in our model.
Experiments verify our enhanced reliability and efficiency.
To the best of our knowledge, this is the first variational  inference modelling of Transformers with the asymmetry in attention kernel addressed.

\section*{Acknowledgements}
This work is jointly supported by the European Research Council under the European Union’s Horizon 2020 research and innovation program/ERC Advanced Grant E-DUALITY (787960), iBOF project Tensor Tools for Taming the Curse (3E221427), Research Council KU Leuven: Optimization framework for deep kernel machines C14/18/068, KU Leuven Grant CoE PFV/10/002, The Research Foundation–Flanders (FWO) projects: GOA4917N (Deep Restricted kernel Machines: Methods and Foundations), Ph.D./Postdoctoral grant, the Flemish Government (AI Research Program), EU H2020 ICT-48 Network TAILOR (Foundations of Trustworthy AI-Integrating Reasoning, Learning and Optimization), Leuven.AI Institute.

\section*{Impact Statement}
In this work, we provide a new variational inference modelling for realizing uncertainty-aware self-attention modules in Transformers relying on Sparse Variational Gaussian Processes and Kernel SVD.
Compared to methods in literature for self-attention within the Bayesian inference framework, our method has a low time complexity, hence faster computation processes.
Therefore, we provide a more energy friendly method whose advantages lie in requiring less power consumption during model training.
Moreover, as we propose a new self-attention mechanism, it can be adopted in many deep learning systems with Transformers as backbones.
However, we should be aware that an undue trust in deep learning models when applying to real-life scenarios can lead to unexpected consequences.
Considering that our uncertainty-aware attention maintains good robustness in failure prediction analysis, distribution-shift and out-of-distribution benchmarks,
we still see opportunities in anticipating and mitigating the risks due to real-world uncertainty during applications in advance.

\bibliography{references} 

\begin{thebibliography}{67}
\providecommand{\natexlab}[1]{#1}
\providecommand{\url}[1]{\texttt{#1}}
\expandafter\ifx\csname urlstyle\endcsname\relax
  \providecommand{\doi}[1]{doi: #1}\else
  \providecommand{\doi}{doi: \begingroup \urlstyle{rm}\Url}\fi

\bibitem[Aitchison et~al.(2021)Aitchison, Yang, and Ober]{aitchison2021deep}
Aitchison, L., Yang, A., and Ober, S.~W.
\newblock Deep kernel processes.
\newblock In \emph{International Conference on Machine Learning}, pp.\
  130--140, 2021.

\bibitem[Blundell et~al.(2015)Blundell, Cornebise, Kavukcuoglu, and
  Wierstra]{blundell2015weight}
Blundell, C., Cornebise, J., Kavukcuoglu, K., and Wierstra, D.
\newblock Weight uncertainty in neural network.
\newblock In \emph{International Conference on Machine Learning}, pp.\
  1613--1622, 2015.

\bibitem[Brier(1950)]{brier1950verification}
Brier, G.~W.
\newblock Verification of forecasts expressed in terms of probability.
\newblock \emph{Monthly Weather Review}, 78\penalty0 (1):\penalty0 1--3, 1950.

\bibitem[Brown et~al.(2020)Brown, Mann, Ryder, Subbiah, Kaplan, Dhariwal,
  Neelakantan, Shyam, Sastry, Askell, et~al.]{brown2020language}
Brown, T., Mann, B., Ryder, N., Subbiah, M., Kaplan, J.~D., Dhariwal, P.,
  Neelakantan, A., Shyam, P., Sastry, G., Askell, A., et~al.
\newblock Language models are few-shot learners.
\newblock \emph{Advances in Neural Information Processing Systems},
  33:\penalty0 1877--1901, 2020.

\bibitem[Chen \& Li(2023)Chen and Li]{chen2023calibrating}
Chen, W. and Li, Y.
\newblock Calibrating transformers via sparse {G}aussian processes.
\newblock In \emph{International Conference on Learning Representations}, 2023.

\bibitem[Chen et~al.(2023)Chen, Tao, Tonin, and Suykens]{chen2023primal}
Chen, Y., Tao, Q., Tonin, F., and Suykens, J.~A.
\newblock Primal-attention: Self-attention through asymmetric kernel {SVD} in
  primal representation.
\newblock \emph{Advances in Neural Information Processing Systems}, 2023.

\bibitem[Chi et~al.(2022)Chi, Fan, Ramadge, and Rudnicky]{chi2022kerple}
Chi, T.-C., Fan, T.-H., Ramadge, P.~J., and Rudnicky, A.
\newblock Kerple: Kernelized relative positional embedding for length
  extrapolation.
\newblock \emph{Advances in Neural Information Processing Systems},
  35:\penalty0 8386--8399, 2022.

\bibitem[Choromanski et~al.(2021)Choromanski, Likhosherstov, Dohan, Song, Gane,
  Sarlos, Hawkins, Davis, Mohiuddin, Kaiser, Belanger, Colwell, and
  Weller]{choromanski2021rethinking}
Choromanski, K.~M., Likhosherstov, V., Dohan, D., Song, X., Gane, A., Sarlos,
  T., Hawkins, P., Davis, J.~Q., Mohiuddin, A., Kaiser, L., Belanger, D.~B.,
  Colwell, L.~J., and Weller, A.
\newblock Rethinking attention with performers.
\newblock In \emph{International Conference on Learning Representations}, 2021.

\bibitem[Cinquin et~al.(2022)Cinquin, Immer, Horn, and
  Fortuin]{cinquin2022pathologies}
Cinquin, T., Immer, A., Horn, M., and Fortuin, V.
\newblock Pathologies in priors and inference for {B}ayesian transformers.
\newblock In \emph{the Fourth Symposium on Advances in Approximate Bayesian
  Inference}, 2022.

\bibitem[Coker et~al.(2022)Coker, Bruinsma, Burt, Pan, and
  Doshi-Velez]{Coker2022WideMeanFieldBayesian}
Coker, B., Bruinsma, W.~P., Burt, D.~R., Pan, W., and Doshi-Velez, F.
\newblock Wide mean-field {B}ayesian neural networks ignore the data.
\newblock In \emph{International Conference on Artificial Intelligence and
  Statistics}, 2022.

\bibitem[Damianou \& Lawrence(2013)Damianou and Lawrence]{damianou2013deep}
Damianou, A. and Lawrence, N.~D.
\newblock Deep {G}aussian processes.
\newblock In \emph{International Conference on Artificial Intelligence and
  Statistics}, pp.\  207--215, 2013.

\bibitem[Davis \& Goadrich(2006)Davis and Goadrich]{davis2006relationship}
Davis, J. and Goadrich, M.
\newblock The relationship between precision-recall and {ROC} curves.
\newblock In \emph{International Conference on Machine Learning}, pp.\
  233--240, 2006.

\bibitem[Dosovitskiy et~al.(2021)Dosovitskiy, Beyer, Kolesnikov, Weissenborn,
  Zhai, Unterthiner, Dehghani, Minderer, Heigold, Gelly, Uszkoreit, and
  Houlsby]{dosovitskiy2021an}
Dosovitskiy, A., Beyer, L., Kolesnikov, A., Weissenborn, D., Zhai, X.,
  Unterthiner, T., Dehghani, M., Minderer, M., Heigold, G., Gelly, S.,
  Uszkoreit, J., and Houlsby, N.
\newblock An image is worth 16x16 words: Transformers for image recognition at
  scale.
\newblock In \emph{International Conference on Learning Representations}, 2021.

\bibitem[Eckart \& Young(1936)Eckart and Young]{eckart1936}
Eckart, C. and Young, G.
\newblock The approximation of one matrix by another of lower rank.
\newblock \emph{Psychometrika}, 1\penalty0 (3):\penalty0 211--218, 1936.

\bibitem[Fan et~al.(2020)Fan, Zhang, Chen, and Zhou]{fan2020bayesian}
Fan, X., Zhang, S., Chen, B., and Zhou, M.
\newblock Bayesian attention modules.
\newblock \emph{Advances in Neural Information Processing Systems},
  33:\penalty0 16362--16376, 2020.

\bibitem[Foong et~al.(2020)Foong, Burt, Li, and
  Turner]{foong2020expressiveness}
Foong, A., Burt, D., Li, Y., and Turner, R.
\newblock On the expressiveness of approximate inference in {B}ayesian neural
  networks.
\newblock \emph{Advances in Neural Information Processing Systems},
  33:\penalty0 15897--15908, 2020.

\bibitem[Gal \& Ghahramani(2016)Gal and Ghahramani]{gal2016dropout}
Gal, Y. and Ghahramani, Z.
\newblock Dropout as a {B}ayesian approximation: Representing model uncertainty
  in deep learning.
\newblock In \emph{International Conference on Machine Learning}, 2016.

\bibitem[Geifman \& El-Yaniv(2017)Geifman and El-Yaniv]{geifman2017selective}
Geifman, Y. and El-Yaniv, R.
\newblock Selective classification for deep neural networks.
\newblock \emph{Advances in Neural Information Processing Systems}, 30, 2017.

\bibitem[Geifman et~al.(2019)Geifman, Uziel, and El-Yaniv]{geifman2019bias}
Geifman, Y., Uziel, G., and El-Yaniv, R.
\newblock Bias-reduced uncertainty estimation for deep neural classifiers.
\newblock \emph{International Conference on Learning Representations}, 2019.

\bibitem[Graves(2011)]{graves2011practical}
Graves, A.
\newblock Practical variational inference for neural networks.
\newblock \emph{Advances in Neural Information Processing Systems}, 24, 2011.

\bibitem[Guo et~al.(2017)Guo, Pleiss, Sun, and Weinberger]{guo2017calibration}
Guo, C., Pleiss, G., Sun, Y., and Weinberger, K.~Q.
\newblock On calibration of modern neural networks.
\newblock In \emph{International Conference on Machine Learning}, pp.\
  1321--1330. PMLR, 2017.

\bibitem[Hendrycks \& Dietterich(2019)Hendrycks and
  Dietterich]{hendrycks2019benchmarking}
Hendrycks, D. and Dietterich, T.
\newblock Benchmarking neural network robustness to common corruptions and
  perturbations.
\newblock \emph{International Conference on Learning Representations}, 2019.

\bibitem[Hendrycks \& Gimpel(2017)Hendrycks and Gimpel]{hendrycks2016baseline}
Hendrycks, D. and Gimpel, K.
\newblock A baseline for detecting misclassified and out-of-distribution
  examples in neural networks.
\newblock In \emph{International Conference on Learning Representations}, 2017.

\bibitem[Kendall \& Gal(2017)Kendall and Gal]{kendall2017uncertainties}
Kendall, A. and Gal, Y.
\newblock What uncertainties do we need in {B}ayesian deep learning for
  computer vision?
\newblock \emph{Advances in Neural Information Processing Systems}, 30, 2017.

\bibitem[Kingma \& Ba(2015)Kingma and Ba]{kingma2014adam}
Kingma, D.~P. and Ba, J.
\newblock Adam: A method for stochastic optimization.
\newblock \emph{International Conference on Learning Representations}, 2015.

\bibitem[Kingma \& Welling(2013)Kingma and Welling]{kingma2013auto}
Kingma, D.~P. and Welling, M.
\newblock Auto-encoding variational bayes.
\newblock \emph{arXiv preprint arXiv:1312.6114}, 2013.

\bibitem[Kristiadi et~al.(2020)Kristiadi, Hein, and Hennig]{kristiadi2020being}
Kristiadi, A., Hein, M., and Hennig, P.
\newblock Being {B}ayesian, even just a bit, fixes overconfidence in {R}e{LU}
  networks.
\newblock In \emph{International Conference on Machine Learning}, pp.\
  5436--5446, 2020.

\bibitem[Krizhevsky et~al.(2009)Krizhevsky, Hinton,
  et~al.]{krizhevsky2009learning}
Krizhevsky, A., Hinton, G., et~al.
\newblock Learning multiple layers of features from tiny images.
\newblock \emph{Technical Report}, 2009.

\bibitem[Lakshminarayanan et~al.(2017)Lakshminarayanan, Pritzel, and
  Blundell]{lakshminarayanan2017simple}
Lakshminarayanan, B., Pritzel, A., and Blundell, C.
\newblock Simple and scalable predictive uncertainty estimation using deep
  ensembles.
\newblock \emph{Advances in Neural Information Processing Systems}, 30, 2017.

\bibitem[Lanczos(1958)]{lanczos1958linear}
Lanczos, C.
\newblock Linear systems in self-adjoint form.
\newblock \emph{The American Mathematical Monthly}, 65\penalty0 (9):\penalty0
  665--679, 1958.

\bibitem[L{\'a}zaro-Gredilla \& Figueiras-Vidal(2009)L{\'a}zaro-Gredilla and
  Figueiras-Vidal]{lazaro2009inter}
L{\'a}zaro-Gredilla, M. and Figueiras-Vidal, A.
\newblock Inter-domain {G}aussian processes for sparse inference using inducing
  features.
\newblock \emph{Advances in Neural Information Processing Systems}, 22, 2009.

\bibitem[Leibfried et~al.(2020)Leibfried, Dutordoir, John, and
  Durrande]{leibfried2020tutorial}
Leibfried, F., Dutordoir, V., John, S., and Durrande, N.
\newblock A tutorial on sparse {G}aussian processes and variational inference.
\newblock \emph{arXiv preprint arXiv:2012.13962}, 2020.

\bibitem[Li et~al.(2024)Li, Chen, Yu, Chen, and Shen]{li2024sure}
Li, Y., Chen, Y., Yu, X., Chen, D., and Shen, X.
\newblock {SURE}: {SU}rvey {RE}cipes for building reliable and robust deep
  networks.
\newblock In \emph{IEEE/CVF Conference on Computer Vision and Pattern
  Recognition}, 2024.

\bibitem[Liu et~al.(2020)Liu, Lin, Padhy, Tran, Bedrax~Weiss, and
  Lakshminarayanan]{liu2020simple}
Liu, J., Lin, Z., Padhy, S., Tran, D., Bedrax~Weiss, T., and Lakshminarayanan,
  B.
\newblock Simple and principled uncertainty estimation with deterministic deep
  learning via distance awareness.
\newblock \emph{Advances in Neural Information Processing Systems},
  33:\penalty0 7498--7512, 2020.

\bibitem[Maas et~al.(2011)Maas, Daly, Pham, Huang, Ng, and
  Potts]{maas2011learning}
Maas, A., Daly, R.~E., Pham, P.~T., Huang, D., Ng, A.~Y., and Potts, C.
\newblock Learning word vectors for sentiment analysis.
\newblock In \emph{Annual Meeting of the Association for Computational
  Linguistics: Human Language Technologies}, pp.\  142--150, 2011.

\bibitem[Matthews(1975)]{matthews1975comparison}
Matthews, B.~W.
\newblock Comparison of the predicted and observed secondary structure of {T}4
  phage lysozyme.
\newblock \emph{Biochimica et Biophysica Acta (BBA)-Protein Structure},
  405\penalty0 (2):\penalty0 442--451, 1975.

\bibitem[Milsom et~al.(2024)Milsom, Anson, and
  Aitchison]{milsom2023convolutional}
Milsom, E., Anson, B., and Aitchison, L.
\newblock Convolutional deep kernel machines.
\newblock \emph{International Conference on Learning Representations}, 2024.

\bibitem[Moon et~al.(2020)Moon, Kim, Shin, and Hwang]{moon2020confidence}
Moon, J., Kim, J., Shin, Y., and Hwang, S.
\newblock Confidence-aware learning for deep neural networks.
\newblock In \emph{International Conference on Machine Learning}, pp.\
  7034--7044, 2020.

\bibitem[Mukhoti et~al.(2020)Mukhoti, Kulharia, Sanyal, Golodetz, Torr, and
  Dokania]{mukhoti2020calibrating}
Mukhoti, J., Kulharia, V., Sanyal, A., Golodetz, S., Torr, P., and Dokania, P.
\newblock Calibrating deep neural networks using focal loss.
\newblock \emph{Advances in Neural Information Processing Systems},
  33:\penalty0 15288--15299, 2020.

\bibitem[Naeini et~al.(2015)Naeini, Cooper, and
  Hauskrecht]{naeini2015obtaining}
Naeini, M.~P., Cooper, G., and Hauskrecht, M.
\newblock Obtaining well calibrated probabilities using {B}ayesian binning.
\newblock In \emph{AAAI Conference on Artificial Intelligence}, volume~29,
  2015.

\bibitem[Netzer et~al.(2011)Netzer, Wang, Coates, Bissacco, Wu, and
  Ng]{netzer2011reading}
Netzer, Y., Wang, T., Coates, A., Bissacco, A., Wu, B., and Ng, A.~Y.
\newblock Reading digits in natural images with unsupervised feature learning.
\newblock 2011.

\bibitem[Nguyen et~al.(2022)Nguyen, Pham, Nguyen, Nguyen, Osher, and
  Ho]{nguyen2022fourierformer}
Nguyen, T., Pham, M., Nguyen, T., Nguyen, K., Osher, S., and Ho, N.
\newblock Fourierformer: Transformer meets generalized fourier integral
  theorem.
\newblock \emph{Advances in Neural Information Processing Systems},
  35:\penalty0 29319--29335, 2022.

\bibitem[Nguyen et~al.(2023)Nguyen, Nguyen, Ho, Bertozzi, Baraniuk, and
  Osher]{nguyen2023a}
Nguyen, T.~M., Nguyen, T.~M., Ho, N., Bertozzi, A.~L., Baraniuk, R., and Osher,
  S.
\newblock A primal-dual framework for transformers and neural networks.
\newblock In \emph{International Conference on Learning Representations}, 2023.

\bibitem[Ober et~al.(2021)Ober, Rasmussen, and van~der Wilk]{ober2021promises}
Ober, S.~W., Rasmussen, C.~E., and van~der Wilk, M.
\newblock The promises and pitfalls of deep kernel learning.
\newblock In \emph{Uncertainty in Artificial Intelligence}, pp.\  1206--1216,
  2021.

\bibitem[Peters et~al.(2018)Peters, Neumann, Iyyer, Gardner, Clark, Lee, and
  Zettlemoyer]{peters2018deep}
Peters, M., Neumann, M., Iyyer, M., Gardner, M., Clark, C., Lee, K., and
  Zettlemoyer, L.
\newblock Deep contextualized word representations.
\newblock \emph{Conference of the North American Chapter of the Association for
  Computational Linguistics}, 2018.

\bibitem[Rasmussen \& Williams(2006)Rasmussen and
  Williams]{rasmussen2006gaussian}
Rasmussen, C.~E. and Williams, C.~K.
\newblock \emph{{G}aussian processes for machine learning}.
\newblock Springer, 2006.

\bibitem[Ritter et~al.(2021)Ritter, Kukla, Zhang, and Li]{ritter2021sparse}
Ritter, H., Kukla, M., Zhang, C., and Li, Y.
\newblock Sparse uncertainty representation in deep learning with inducing
  weights.
\newblock \emph{Advances in Neural Information Processing Systems},
  34:\penalty0 6515--6528, 2021.

\bibitem[Salimbeni \& Deisenroth(2017)Salimbeni and
  Deisenroth]{salimbeni2017doubly}
Salimbeni, H. and Deisenroth, M.
\newblock Doubly stochastic variational inference for deep {G}aussian
  processes.
\newblock \emph{Advances in Neural Information Processing Systems}, 30, 2017.

\bibitem[Schmidt(1907)]{schmidt1907theorie}
Schmidt, E.
\newblock Zur theorie der linearen und nichtlinearen integralgleichungen.
\newblock \emph{Mathematische Annalen}, 63\penalty0 (4):\penalty0 433--476,
  1907.

\bibitem[Stewart(1993)]{stewart1993early}
Stewart, G.~W.
\newblock On the early history of the singular value decomposition.
\newblock \emph{SIAM Review}, 35\penalty0 (4):\penalty0 551--566, 1993.

\bibitem[Suykens(2016)]{suykens2016svd}
Suykens, J.~A.
\newblock {SVD} revisited: A new variational principle, compatible feature maps
  and nonlinear extensions.
\newblock \emph{Applied and Computational Harmonic Analysis}, 40\penalty0
  (3):\penalty0 600--609, 2016.

\bibitem[Tao et~al.(2023)Tao, Tonin, Patrinos, and Suykens]{tao2023nonlinear}
Tao, Q., Tonin, F., Patrinos, P., and Suykens, J.~A.
\newblock Nonlinear {SVD} with asymmetric kernels: feature learning and
  asymmetric {N}ystr\"om method.
\newblock \emph{arXiv preprint arXiv:2306.07040}, 2023.

\bibitem[Titsias(2009)]{titsias2009variational}
Titsias, M.
\newblock Variational learning of inducing variables in sparse {G}aussian
  processes.
\newblock In \emph{International Conference on Artificial Intelligence and
  Statistics}, pp.\  567--574, 2009.

\bibitem[Touvron et~al.(2021)Touvron, Cord, Douze, Massa, Sablayrolles, and
  J{\'e}gou]{touvron2021training}
Touvron, H., Cord, M., Douze, M., Massa, F., Sablayrolles, A., and J{\'e}gou,
  H.
\newblock Training data-efficient image transformers \& distillation through
  attention.
\newblock In \emph{International Conference on Machine Learning}, pp.\
  10347--10357, 2021.

\bibitem[Tran et~al.(2019)Tran, Dusenberry, Van Der~Wilk, and
  Hafner]{tran2019bayesian}
Tran, D., Dusenberry, M., Van Der~Wilk, M., and Hafner, D.
\newblock Bayesian layers: A module for neural network uncertainty.
\newblock \emph{Advances in Neural Information Processing Systems}, 32, 2019.

\bibitem[Tsai et~al.(2019)Tsai, Bai, Yamada, Morency, and
  Salakhutdinov]{tsai2019transformer}
Tsai, Y.-H.~H., Bai, S., Yamada, M., Morency, L.-P., and Salakhutdinov, R.
\newblock Transformer dissection: An unified understanding for transformer’s
  attention via the lens of kernel.
\newblock In \emph{Conference on Empirical Methods in Natural Language
  Processing and the International Joint Conference on Natural Language
  Processing (EMNLP-IJCNLP)}, pp.\  4344--4353, 2019.

\bibitem[Vaswani et~al.(2017)Vaswani, Shazeer, Parmar, Uszkoreit, Jones, Gomez,
  Kaiser, and Polosukhin]{vaswani2017attention}
Vaswani, A., Shazeer, N., Parmar, N., Uszkoreit, J., Jones, L., Gomez, A.~N.,
  Kaiser, {\L}., and Polosukhin, I.
\newblock Attention is all you need.
\newblock \emph{Advances in Neural Information Processing Systems}, 30, 2017.

\bibitem[Wang et~al.(2020)Wang, Li, Khabsa, Fang, and Ma]{wang2020linformer}
Wang, S., Li, B.~Z., Khabsa, M., Fang, H., and Ma, H.
\newblock Linformer: Self-attention with linear complexity.
\newblock \emph{arXiv preprint arXiv:2006.04768}, 2020.

\bibitem[Warstadt et~al.(2019)Warstadt, Singh, and Bowman]{warstadt2019neural}
Warstadt, A., Singh, A., and Bowman, S.~R.
\newblock Neural network acceptability judgments.
\newblock \emph{Transactions of the Association for Computational Linguistics},
  7:\penalty0 625--641, 2019.

\bibitem[Williams \& Seeger(2000)Williams and Seeger]{NIPS2000_19de10ad}
Williams, C. and Seeger, M.
\newblock Using the {N}ystr\"{o}m method to speed up kernel machines.
\newblock In Leen, T., Dietterich, T., and Tresp, V. (eds.), \emph{Advances in
  Neural Information Processing Systems}, volume~13, 2000.

\bibitem[Wilson et~al.(2016)Wilson, Hu, Salakhutdinov, and
  Xing]{wilson2016stochastic}
Wilson, A.~G., Hu, Z., Salakhutdinov, R.~R., and Xing, E.~P.
\newblock Stochastic variational deep kernel learning.
\newblock \emph{Advances in Neural Information Processing Systems}, 29, 2016.

\bibitem[Wright \& Gonzalez(2021)Wright and Gonzalez]{wright2021transformers}
Wright, M.~A. and Gonzalez, J.~E.
\newblock Transformers are deep infinite-dimensional non-{M}ercer binary kernel
  machines.
\newblock \emph{arXiv preprint arXiv:2106.01506}, 2021.

\bibitem[Wu et~al.(2022)Wu, Wu, Xu, Wang, and Long]{wu2022flowformer}
Wu, H., Wu, J., Xu, J., Wang, J., and Long, M.
\newblock Flowformer: Linearizing transformers with conservation flows.
\newblock In \emph{International Conference on Machine Learning}, pp.\
  24226--24242, 2022.

\bibitem[Xue et~al.(2021)Xue, Yu, Xu, Liu, Hu, Ye, Geng, Liu, and
  Meng]{xue2021bayesian}
Xue, B., Yu, J., Xu, J., Liu, S., Hu, S., Ye, Z., Geng, M., Liu, X., and Meng,
  H.
\newblock Bayesian transformer language models for speech recognition.
\newblock In \emph{IEEE International Conference on Acoustics, Speech and
  Signal Processing}, pp.\  7378--7382. IEEE, 2021.

\bibitem[Yu et~al.(2015)Yu, Seff, Zhang, Song, Funkhouser, and
  Xiao]{yu2015lsun}
Yu, F., Seff, A., Zhang, Y., Song, S., Funkhouser, T., and Xiao, J.
\newblock Lsun: Construction of a large-scale image dataset using deep learning
  with humans in the loop.
\newblock \emph{arXiv preprint arXiv:1506.03365}, 2015.

\bibitem[Zhang et~al.(2020)Zhang, Li, Zhang, Chen, and
  Wilson]{Zhang2020Cyclical}
Zhang, R., Li, C., Zhang, J., Chen, C., and Wilson, A.~G.
\newblock Cyclical stochastic gradient {MCMC} for {B}ayesian deep learning.
\newblock In \emph{International Conference on Learning Representations}, 2020.

\bibitem[Zhu et~al.(2023)Zhu, Cheng, Zhang, and Liu]{zhu2023openmix}
Zhu, F., Cheng, Z., Zhang, X.-Y., and Liu, C.-L.
\newblock Openmix: Exploring outlier samples for misclassification detection.
\newblock In \emph{IEEE/CVF Conference on Computer Vision and Pattern
  Recognition}, pp.\  12074--12083, 2023.

\end{thebibliography}
\bibliographystyle{icml2024}

\newpage
\appendix
\onecolumn
\section{{Analytical Derivations of SVGPs}}
\subsection{Derivations of \eqref{eq::approx_posterior}: Variational Marginal Distribution in SVGPs}\label{sec:appendix:svgp:vanilla}
As an efficient alternative to the classical GPs \eqref{eq::posterior_vanilla}, SVGPs \cite{titsias2009variational} variationally approximate the Gaussian posterior distribution with a small set of $s$ supports, i.e., 
$(Z, \bm{u}):=\{(\bm{z}_m, u_m)\}_{m=1}^s$, $\bm{z}_m\in \mathcal{X}$, $u_m=f(\bm{z}_m)\in\mathbb{R}$, commonly with $s \ll N$, 
where $Z$ are named ``inducing points'' and $\bm{u}$ are ``inducing variables''.
In SVGPs, the mean $\bm{\mu}_{\bm{u}}$ is often set to zero and the covariance matrix is given by $K_{\bm{u}\bm{u}}:=[\kappa(\bm{z}_i,\bm{z}_j)]\in\mathbb{R}^{s\times s}$. 
The joint GP and the conditional GP of $f(\cdot)$ conditioned on $\bm{u}$ are as follows:

\begin{equation}
     \label{eq::jointGP}
     \begin{array}{c}
   \text{Prior:}\,
    \begin{pmatrix}
        f(\cdot) \\ \bm{u}
    \end{pmatrix}
    \sim \mathcal{GP}
    \left( \bm{0},
    \begin{bmatrix}
        \kappa(\cdot,\cdot) & \bm{\kappa}_{\cdot\bm{u}}
        \\
       \bm{\kappa}_{\bm{u}\cdot} & K_{\bm{u}\bm{u}}
    \end{bmatrix}
    \right)
    \vspace{0.15cm}
    \\
    \Downarrow
    \\
    f(\cdot)|\bm{u}\sim \mathcal{GP}
    \left(\bm{\kappa}_{\cdot\bm{u}}K_{\bm{uu}}^{-1}\bm{u},\,
    \kappa(\cdot,\cdot) - \bm{\kappa}_{\cdot\bm{u}} K_{\bm{uu}}^{-1}\bm{\kappa}_{\bm{u}\cdot}
    \right)
    \end{array}
\end{equation}
with
$\bm{\kappa}_{\cdot\bm{u}}:=[\kappa(\cdot,\bm{z}_1),\ldots,\kappa(\cdot,\bm{z}_s)]$, 
$\bm{\kappa}_{\bm{u}\cdot}:=[\kappa(\bm{z}_1,\cdot),\ldots,\kappa(\bm{z}_s,\cdot)]^\top$, and
$\bm{\kappa}_{\bm{u}\cdot}=\bm{\kappa}_{\cdot\bm{u}}^\top$.
Then, the corresponding conditional distribution of the function values $\bm{f}$ on $\bm{u}$ is as follows:
\begin{equation*}
    \mathrm{p}(\bm{f}|\bm{u})=\mathcal{N}(K_{XZ}K_{ZZ}^{-1}\bm{u},\,K_{XX}-K_{XZ}K_{ZZ}^{-1}K_{ZX}),
\end{equation*}
where $K_{XZ}:=[\kappa(\bm{x}_i,\bm{z}_j)]\in\mathbb{R}^{N\times s}$, 
$K_{ZX}:=[\kappa(\bm{z}_i, \bm{x}_j)]\in\mathbb{R}^{s\times N}$
and $K_{ZZ}=K_{\bm{u}\bm{u}}\in\mathbb{R}^{s\times s}$.

In addition to considering the marginal distribution $\mathrm{p}(\bm{u})=\mathcal{N}(\bm{0},K_{ZZ})$, SVGPs provide a variational distribution $\mathrm{q}(\bm{u})=\mathcal{N}(\bm{m}_{\bm{u}}, S_{\bm{u}\bm{u}})$ where $\bm{m}_{\bm{u}}\in\mathbb{R}^{s}$, $S_{\bm{uu}}\in\mathbb{R}^{s\times s}$ \cite{leibfried2020tutorial}.
The variational marginal distribution of $\bm{f}$ is given  by $\mathrm{q}(\bm{f})=\int \mathrm{p}(\bm{f}|\bm{u})\mathrm{q}(\bm{u})\,\mathrm{d}\bm{u}$, which is still Gaussian and corresponds to the approximate posterior \eqref{eq::approx_posterior} in Section \ref{subsec::SVGP}:
\begin{equation} \label{eq::apdx::can_pos}
    \mathrm{q}(\bm{f})= \mathcal{N}\left(K_{XZ}K_{ZZ}^{-1}\bm{m}_{\bm{u}},\, 
    K_{XX}-K_{XZ}K_{ZZ}^{-1}(K_{ZZ}-S_{\bm{u}\bm{u}})K_{ZZ}^{-1}K_{ZX}\right).
\end{equation}

In inference, the posterior distribution of $\bm{f}^*$ evaluated at test inputs $X^*$ 
is then given by
\begin{equation*} 
    \mathrm{q}(\bm{f}^*|X^*,Z)
    =\mathcal{N}\left(K_{X^*Z}K_{ZZ}^{-1}\bm{m}_{\bm{u}},\, 
     K_{X^*X^*}-K_{X^*Z}K_{ZZ}^{-1}(K_{ZZ}-S_{\bm{u}\bm{u}})K_{ZZ}^{-1}K_{ZX^*}\right).
\end{equation*}

In SVGPs, the evidence lower-bound (ELBO) objective involves the variational parameters $\bm{m}_{\bm{u}}$, $S_{\bm{uu}}$ in the variational distribution $\mathrm{q}(\bm{u})=\mathcal{N}(\bm{m}_{\bm{u}}, S_{\bm{u}\bm{u}})$ and is used for the training
\begin{equation*}
    \mathcal{L}_{\text{ELBO}}=\mathbb{E}_{\mathrm{q}(\bm{f})}\left[\log \mathrm{p}(\bm{y}|\bm{f})\right] - \text{KL}\left(\mathrm{q}(\bm{u})|| \mathrm{p}(\bm{u})\right).
\end{equation*}
The derivation of this classical ELBO is provided in \eqref{eq::dev_elbo} in Appendix \ref{sec:elbo:appendix}.
More details on SVGPs can refer to \citet{titsias2009variational,leibfried2020tutorial}.

\subsection{Derivations of \eqref{eq::approx_pos_sparse}: SVGPs with Kernel-Eigen Features}\label{sec:appendix:svgp:kernel_eigen}
With \eqref{eq:inducing:features}, let $\phi_m(\cdot):=\nu_m(\cdot)$ be chosen as the eigenfunction corresponding to the $m$-th largest eigenvalue $\lambda_m$ w.r.t. the kernel function $\kappa(\cdot,\cdot)$, i.e., 
$\int \kappa(\cdot,\bm{x})\nu_m(\bm{x})\,\mathrm{d}\bm{x}=\lambda_m \nu_m(\cdot)$ in \eqref{eq:evd:integral}.
In this setup, the $\bm{\kappa}_{\cdot\bm{u}}$, $\bm{\kappa}_{\bm{u}\cdot}$, $K_{\bm{u}\bm{u}}$ of the prior GP in \eqref{eq::jointGP} are updated as follows \cite{leibfried2020tutorial}:
\begin{equation}
    \label{eq::eigen_gp}
    \begin{array}{c}
    \bm{\kappa}_{\cdot\bm{u}}[m]
    =\bm{\kappa}_{\bm{u}\cdot}[m]
    =\lambda_m \nu_m(\cdot), 
    \\
    K_{\bm{u}\bm{u}}=\text{diag}\{\lambda_1,\ldots,\lambda_s\},
        \end{array}
\end{equation}
where $\bm{\kappa}_{\cdot\bm{u}}[m]$, $\bm{\kappa}_{\bm{u}\cdot}[m]$ are the $m$-th entry of $\bm{\kappa}_{\cdot\bm{u}}$ and $\bm{\kappa}_{\bm{u}\cdot}$ respectively.
The derivations of \eqref{eq::eigen_gp} are provided in the following.

The cross-covariance $\bm{\kappa}_{\cdot\bm{u}}$ is a vector-valued function with $s$ outputs.
The scalar-valued function $\bm{\kappa}_{\cdot\bm{u}}[m]$ corresponding to the output index $m$ is computed as:
\begin{equation*}
\begin{split}
    \bm{\kappa}_{\cdot\bm{u}}[m]
    &=\mathbb{E}\left[\left(f(\cdot)-0\right)\left(u_m-0\right)\right]
    = \mathbb{E}\left[f(\cdot)\left(\int f(\bm{x})\nu_m(\bm{x})\,\mathrm{d}\bm{x} \right)\right]
    \\
    &=\int \mathbb{E}\left[f(\cdot)f(\bm{x})\right]\nu_m(\bm{x})\,\mathrm{d}\bm{x}
    = \int \kappa(\cdot,\bm{x})\nu_m(\bm{x})\,\mathrm{d}\bm{x} 
    =\lambda_m\nu_m(\cdot).
\end{split}
\end{equation*}
Similarly, the $m$-th output of the cross-covariance $\bm{\kappa}_{\bm{u}\cdot}$ is the attained as:
\begin{equation*}
\begin{split}
    \bm{\kappa}_{\bm{u}\cdot}[m]
    &=\mathbb{E}\left[\left(u_m-0\right)\left(f(\cdot)-0\right)\right]
    = \mathbb{E}\left[\left(\int f(\bm{x})\nu_m(\bm{x})\,\mathrm{d}\bm{x} \right)f(\cdot)\right]
    \\
    &=\int \mathbb{E}\left[f(\bm{x})f(\cdot)\right]\nu_m(\bm{x})\,\mathrm{d}\bm{x}
    =\int \kappa(\bm{x},\cdot)\nu_m(\bm{x})\,\mathrm{d}\bm{x} 
    \\
    &=\int \kappa(\cdot,\bm{x})\nu_m(\bm{x})\,\mathrm{d}\bm{x} 
    =\lambda_m\nu_m(\cdot),
\end{split}
\end{equation*}
where the {fifth equation} holds with a symmetric $\kappa(\cdot,\cdot)$.
The covariance $K_{\bm{u}\bm{u}}$ is an $s\times s$ matrix with the $[i,j]$-th entry computed by
\begin{equation}\label{eq:appendix:kuu}
\begin{split}
    K_{\bm{u}\bm{u}}[i,j]
    &=\mathbb{E}\left[\left(u_i-0\right)\left(u_j-0\right)\right]
    = \mathbb{E}\left[\left(\int f(\bm{x})\nu_i(\bm{x})\,\mathrm{d}\bm{x} \right)\left(\int f(\bm{x}')\nu_j(\bm{x}')\,\mathrm{d}\bm{x}' \right)\right]
    \\
    &=\int\int \mathbb{E}\left[f(\bm{x})f(\bm{x}')\right]\nu_i(\bm{x})\nu_j(\bm{x}')\,\mathrm{d}\bm{x}'\mathrm{d}\bm{x}
    = \int\int \kappa(\bm{x},\bm{x}')\nu_i(\bm{x})\nu_j(\bm{x}')\,\mathrm{d}\bm{x}'\mathrm{d}\bm{x}
    \\
    & = \int \nu_i(\bm{x}) \int \kappa(\bm{x},\bm{x}')\nu_j(\bm{x}')\,\mathrm{d}\bm{x}'\mathrm{d}\bm{x}
    = \int \nu_i(\bm{x})\lambda_j\nu_j(\bm{x})\mathrm{d}\bm{x}
    = \lambda_j \int \nu_i(\bm{x})\nu_j(\bm{x})\mathrm{d}\bm{x}
    \\
    & = \lambda_i\,
    \text{if $i=j$ else $0$}.
\end{split}
\end{equation}
The last equation in \eqref{eq:appendix:kuu} establishes since eigenfunctions are orthonormal systems: $\int\nu_i(\bm{x})\nu_j(\bm{x})\,\mathrm{d}\bm{x}$ equals one if $i$ equals $j$, and is zero otherwise. 
When evaluating the SVGP prior over a finite set $X\subset\mathcal{X}$ and its inducing points $Z$, we have the finite case of the integral equations w.r.t. the symmetric kernel function $\kappa(\cdot,\cdot)$ in \eqref{eq:evd:integral} \cite{NIPS2000_19de10ad} as
\begin{equation*}
    K_{XX}H=H\Lambda,
\end{equation*}
where $H:=[\bm{\nu}_1,\ldots,\bm{\nu}_s]\in\mathbb{R}^{N\times s}$ contains the eigenvectors to the top-$s$ nonzero eigenvalues of the kernel matrix $K_{XX}$, i.e., $\Lambda=\text{diag}\{\lambda_1,\ldots,\lambda_s\}$.
By substituting \eqref{eq::eigen_gp} together with its finite sample case \eqref{eq:evd:svgp} into the GP prior in \eqref{eq::jointGP}, with $\mathrm{q}(\bm{u})=\mathcal{N}(\bm{m}_{\bm{u}},S_{\bm{uu}})$, the posterior distribution corresponding to \eqref{eq::apdx::can_pos} is then formulated as 
\begin{equation}
    \label{eq:appendix:prior:posterior:svgp}
     \begin{array}{c}
   \text{Prior:}
    \begin{pmatrix}
        \bm{f} \\ \bm{u}
    \end{pmatrix}
    \sim \mathcal{N}
    \left( \bm{0},
    \begin{bmatrix}
        K_{XX} & H\Lambda
        \\
        \Lambda H^\top & \Lambda
    \end{bmatrix}
    \right) 
    \vspace{0.15cm}
    \\
    \Downarrow
    \\
    \mathrm{q}(\bm{f})=\mathcal{N}\left( 
    (H\Lambda)\Lambda^{-1}\bm{m}_{\bm{u}},\,
    K_{XX} - (H\Lambda)\Lambda^{-1}(\Lambda-S_{\bm{u}\bm{u}})\Lambda^{-1}(\Lambda H^\top)
    \right).
    \end{array}
\end{equation}
This completes the derivation of \eqref{eq::approx_pos_sparse} in the paper.

\section{Analytical Derivations of KEP-SVGP}
\subsection{Derivations of \eqref{eq::two_symm_eigen}: Two Eigenvalue Problems Induced by KSVD on the Asymmetric Attention Kernel Matrix}
\label{sec:two:evd:appendix}
From the KSVD on the asymmetric attention kernel matrix $K_{\rm att}$, we have $K_{\rm att}=H_e\Lambda H_r^\top$, where the left and right singular vectors suffice $H_e^\top H_e=I$, $H_r^\top H_r=I$ with $\bm{h}_{e,i}:=H_e[:,i]=[\bm{h}_{e_1}[i],\ldots,\bm{h}_{e_N}[i]]^\top\in\mathbb{R}^N$ and $\bm{h}_{r,i}:=H_r[:,i]=[\bm{h}_{r_1}[i],\ldots,\bm{h}_{r_N}[i]]^\top\in\mathbb{R}^N$, $i=1,\ldots,N$. 
 With KSVD, we have the shifted eigenvalue problem in \eqref{eq::shifted_eigen}, which gives
\begin{equation}
    \begin{array}{rl}
    & \left(K_{\rm att}K_{\rm att}^\top\right) H_e=K_{\rm att}\left(H_e\Lambda H_r^\top\right)^\top H_e=K_{\rm att}H_r\Lambda H_e^\top H_e=K_{\rm att}H_r\Lambda \left(H_e^\top H_e\right)
    =K_{\rm att} H_r\Lambda = H_e \Lambda^2,
    \\
    & \left(K_{\rm att}^\top K_{\rm att}\right) H_r=\left(H_e\Lambda H_r^\top\right)^\top K_{\rm att} H_r=H_r\Lambda H_e^\top \left(K_{\rm att} H_r\right)
    = H_r\Lambda H_e^\top \left(H_e\Lambda\right) = H_r\Lambda \left(H_e^\top H_e\right)\Lambda = H_r\Lambda^2.
\end{array}
\end{equation}
More explanations on the shifted eigenvalue problem from SVD can refer to the Lanczos decomposition in \citet{lanczos1958linear}, Theorem 3.2 in \citet{chen2023primal}, Proposition 3.1 in \citet{tao2023nonlinear}. 
This completes the derivation of \eqref{eq::two_symm_eigen} in the paper.

\subsection{Derivations of \eqref{eq::e:r:score_sgp}: SVGP Pair on the Asymmetric Attention Kernel}
\label{sec:appendix:twin:svgp} 
Within the framework of KSVD \cite{suykens2016svd,tao2023nonlinear} w.r.t. the self-attention \cite{chen2023primal}, we have the equivalence between the primal and dual model representation for the projection matrices w.r.t. right and left singular vectors of KSVD in \eqref{eq::primal_dual}, such that 
\begin{equation}
    E_X:=W_{e}^\top\phi_q(X)
    \overset{\eqref{eq::primal_dual}}{=}
    K_{\rm att}H_r
    \overset{\eqref{eq::shifted_eigen}}{=}
    H_e\Lambda, 
    \quad 
    R_X:=W_{r}^\top\phi_k(X)
    \overset{\eqref{eq::primal_dual}}{=}
    K^\top H_e
    \overset{\eqref{eq::shifted_eigen}}{=}
    H_r\Lambda,
\end{equation}
where $E_X:=e(X) =[e(\bm{x}_i),\ldots,e(\bm{x}_N)]^\top\in\mathbb{R}^{N\times s}$, $R_X:=R_X=[r(\bm{x}_i),\ldots,r(\bm{x}_N)]^\top \in \mathbb R^{N\times s}$ are the projection matrices w.r.t. right and left singular vectors of KSVD in \eqref{eq::primal_dual}.  
Here, we only consider the output of the $d$-th dimension.
We consider the approximate posterior GP w.r.t. the symmetric kernel $K_{\rm att}K_{\rm att}^\top$ and $K_{\rm att}^\top K_{\rm att}$ given the distribution on the inducing points $\bm{u}^e_{[d]}, \bm{u}^r_{[d]}\sim\mathcal{N}(\bm{m}_{\bm{u},[d]},S_{\bm{uu},[d]})$, where 
$\bm{m}_{\bm{u},{[d]}}:=\bm{m}_{\bm{u}}[:,d]\in\mathbb{R}^s$ and
$S_{\bm{uu},[d]}:=S_{\bm{uu}}[:,:,d] \in\mathbb{R}^{s\times s}$  correspond  to the $d$-th output dimension for  the variational parameters $\bm{m}_{\bm{u}}\in\mathbb{R}^{s\times s}$, 
$S_{\bm{uu}}\in\mathbb{R}^{s\times s \times s}$.

According to the formulations of SVGPs in \eqref{eq::approx_pos_sparse} and \eqref{eq:appendix:prior:posterior:svgp}, the mean of the posterior process in the first set of SVGP in \eqref{eq::e:r:score_sgp} is attained as
\begin{align}
    \bm{\mu}^e:=
    (H_e\Lambda^2)\Lambda^{-2}\bm{m}_{\bm{u},[d]}
    = (H_e\Lambda)\Lambda^{-1}\bm{m}_{\bm{u},[d]}
    = E_X\Lambda^{-1}\bm{m}_{\bm{u},[d]},
\end{align}
with the covariance matrix
\begin{align*}
    \Sigma^e
    & :=
    K_{\rm att}K_{\rm att}^\top -(H_e\Lambda^2)\Lambda^{-2}(\Lambda^2-S_{\bm{uu},[d]})\Lambda^{-2}(\Lambda^2H_e^\top)
    \\
    & = K_{\rm att}K_{\rm att}^\top - (H_e\Lambda^2)\Lambda^{-2}(\Lambda^2H_e^\top) + (H_e\Lambda)\Lambda^{-1}S_{\bm{uu},[d]}\Lambda^{-1}(H_e\Lambda)^\top
    \\
    & = \underbrace{K_{\rm att}K_{\rm att}^\top - H_e\Lambda^2H_e^\top}_{\approx 0 \ (\text{see Remark} \ \ref{rmk:approx})} + E_X\Lambda^{-1}S_{\bm{uu},[d]}\Lambda^{-1}E_X^\top
    \\
    & \approx E_X\Lambda^{-1}S_{\bm{uu},[d]}\Lambda^{-1}E_X^\top
    \\
    & = E_X\Lambda^{-2}S_{\bm{uu},[d]} E_X^\top.
\end{align*}

The attention matrix is commonly low-rank \cite{wang2020linformer,chen2023primal}, so we motivate to utilize the fast-to-compute approximate posterior as given by Remark \ref{rmk:approx}. Numerical evidence is also provided in Appendix \ref{appdx::low_rank}, verifying the validity of Remark \ref{rmk:approx} in our work.
Therefore, we have the approximate distribution of the posterior process as
\begin{align*}
   \tilde{\mathrm{q}}(\bm{f}^e_{[d]}) =\mathcal{N}\left(E_X\Lambda^{-1}\bm{m}_{\bm{u},[d]},\, E_X\Lambda^{-2}S_{\bm{uu},[d]} E_X^\top\right).
\end{align*}
Similarly, based on \eqref{eq::e:r:score_sgp} and \eqref{eq::approx_pos_sparse}, the mean of the posterior process w.r.t. the symmetric kernel $K_{\rm att}^\top K_{\rm att}$ is  
\begin{align*}
    \bm{\mu}^r:=
    (H_r\Lambda^2)\Lambda^{-2}\bm{m}_{\bm{u},[d]}
    =(H_r\Lambda)\Lambda^{-1}\bm{m}_{\bm{u},[d]}
    = R_X\Lambda^{-1}\bm{m}_{\bm{u},[d]},
\end{align*} 
as given in \eqref{eq::e:r:score_sgp}, and the corresponding covariance is 
\begin{align*}
    \Sigma^r
    & :=
    K_{\rm att}^\top K_{\rm att} -(H_r\Lambda^2)\Lambda^{-2}(\Lambda^2-S_{\bm{uu},[d]})\Lambda^{-2}(\Lambda^2H_r^\top)
    \\
    & = K_{\rm att}^\top K_{\rm att} - (H_r\Lambda^2)\Lambda^{-2}(\Lambda^2H_r^\top) + (H_r\Lambda)\Lambda^{-1}S_{\bm{uu},[d]}\Lambda^{-1}(H_r\Lambda)^\top
    \\
    & = \underbrace{K_{\rm att}^\top K_{\rm att} - H_r\Lambda^2H_r^\top}_{\approx 0 \ (\text{see Remark} \ \ref{rmk:approx})} + R_X\Lambda^{-1}S_{\bm{uu},[d]}\Lambda^{-1}R_X^\top
    \\
    & {\approx}   
    R_X\Lambda^{-1}S_{\bm{uu},[d]}\Lambda^{-1}R_X^\top
    \\
    & = R_X\Lambda^{-2}S_{\bm{uu},[d]} R_X^\top,
\end{align*} 
hence yielding the approximate distribution of the posterior process as
\begin{align*}
\tilde{\mathrm{q}}(\bm{f}^r_{[d]})=\mathcal{N}\left(R_X\Lambda^{-1}\bm{m}_{\bm{u},[d]},\, R_X\Lambda^{-2}S_{\bm{uu},[d]} R_X^\top\right).
\end{align*} 
This completes the derivation of \eqref{eq::e:r:score_sgp} in the paper.

\subsection{Derivations of \eqref{eq::elbo_spg}: The ELBO Objective of KEP-SVGP}\label{sec:elbo:appendix}
For the optimization of our approximate posterior distribution, we 
follow the spirit of deep Gaussian Processes in \citet{salimbeni2017doubly,damianou2013deep}. 
Hence the Transformers applied with KEP-SVGP can be viewed as a sparse approximation to a deep GPs with 
kernels in each layer. 
To proceed  our derivations,
we firstly recall the formulations of the ELBO involving 
$\bm{y},\bm{f},\bm{u}$:
\begin{align} \label{eq::dev_elbo}
    \log \mathrm{p}(\bm{y},\bm{f},\bm{u})
    & =\log\int
    \underbrace{\mathrm{p}(\bm{y}|\bm{f},\bm{u})}_{\text{likelihood}}
    \underbrace{\mathrm{p}(\bm{f},\bm{u})}_{\text{GP Prior}}
    \,\mathrm{d}\bm{f}\mathrm{d}\bm{u}
    = \log\int \mathrm{q}(\bm{f},\bm{u}) \frac{\mathrm{p}(\bm{y}|\bm{f},\bm{u})\mathrm{p}(\bm{f},\bm{u})}{\mathrm{q}(\bm{f},\bm{u})}\,\mathrm{d}\bm{f}\mathrm{d}\bm{u}
    \nonumber\\
    & \overset{\text{Jensen's inequality}}{\geq} \int \mathrm{q}(\bm{f},\bm{u}) \log \left(\frac{\mathrm{p}(\bm{y}|\bm{f},\bm{u})\mathrm{p}(\bm{f},\bm{u})}{\mathrm{q}(\bm{f},\bm{u})}\right)\,\mathrm{d}\bm{f}\mathrm{d}\bm{u}
    \nonumber\\
    & = \int \mathrm{q}(\bm{f},\bm{u}) \log \left(\frac{\mathrm{p}(\bm{y},\bm{f},\bm{u})}{\mathrm{q}(\bm{f},\bm{u})}\right)\,\mathrm{d}\bm{f}\mathrm{d}\bm{u}
    = \mathbb{E}_{\mathrm{q}(\bm{f},\bm{u})}\left[\log\frac{\mathrm{p}(\bm{y},\bm{f},\bm{u})}{\mathrm{q}(\bm{f},\bm{u})}\right].
\end{align}

Similar to the doubly stochastic variational inference framework \cite{salimbeni2017doubly} for the ELBO derivation,
let $\{F^{l}\in\mathbb{R}^{N\times ({N_{\text{h}}} d_v)}\}_{l=1}^L$ be the output of the $l$-th KEP-SVGP layer, where $L$ is the number of layers and $N_{\text{h}}$ is the number of heads. Our model follows the convention of concatenating these multiple heads in canonical self-attention. 
As is shown in \eqref{eq::e:r:score_sgp} and \eqref{eq::same_var_dis}, $\bm{u}^e$, $\bm{u}^r$ share the same marginal prior and variational distribution, hence we only need to optimize one set of variational parameters $\{\bm{m}_{\bm u}, S_{\bm{uu}}\}$ for each KEP-SVGP layer.
In this manner, we consider $\{\bm{u}^{l, {n_{\text{h}}}}\}_{l=1, {n_{\text{h}}}=1}^{L,{N_{\text{h}}}}$ for the $L$ attention layers with $N_{\text{h}}$ heads, and the resulting process can be characterized with the joint density:
\begin{align} \label{eq::joint_density}
    \mathrm{p}\left(Y,\{F^l\}_{l=1}^L, \{\bm{u}^{l, {n_{\text{h}}}}\}_{l=1, {n_{\text{h}}}=1}^{L, {N_{\text{h}}}}|F^0\right)
    =\underbrace{\mathrm{p}\left(Y|F^L\right)}_{\text{likelihood}}
    \underbrace{\prod\nolimits_{l=1}^L\mathrm{p}\left(F^l| \{\bm{u}^{l, {n_{\text{h}}}}\}_{ {n_{\text{h}}}=1}^{ {N_{\text{h}}}},F^{l-1}\right)\mathrm{p}\left(\{\bm{u}^{l, {n_{\text{h}}}}\}_{ {n_{\text{h}}}=1}^{ {N_{\text{h}}}})|F^{l-1}\right)}_{\text{GP Prior}},
\end{align}
where we define $F^0:=X_{\text{in}}$ as the inputs to the Transformer.
The variational posterior of $(\{F^l\}_{l=1}^L, 
\{\bm{u}^{l, {n_{\text{h}}}}\}_{l=1, {n_{\text{h}}}=1}^{L, {N_{\text{h}}}})$ is then:
\begin{align} \label{eq::var_post}
    \mathrm{q}\left(\{F^l\}_{l=1}^L, \{\bm{u}^{l, {n_{\text{h}}}}\}_{l=1, {n_{\text{h}}}=1}^{L, {N_{\text{h}}}}|F^0\right)
    = \prod\nolimits_{l=1}^{L}\mathrm{p}(F^l|\{\bm{u}^{l, {n_{\text{h}}}}\}_{ {n_{\text{h}}}=1}^{ {N_{\text{h}}}}, F^{l-1})
    \mathrm{q}\left(\{\bm{u}^{l, {n_{\text{h}}}}\}_{ {n_{\text{h}}}=1}^{ {N_{\text{h}}}})|F^{l-1}\right),
\end{align}
where $\mathrm{q}\big(\{\bm{u}^{l, {n_{\text{h}}}}\}_{ {n_{\text{h}}}=1}^{ {N_{\text{h}}}})|F^{l-1}\big)$ is the variational distribution, and 
$\mathrm{q}(F^l|\{\bm{u}^{l, {n_{\text{h}}}}\}_{ {n_{\text{h}}}=1}^{ {N_{\text{h}}}}, F^{l-1})=\mathrm{p}(F^l|\{\bm{u}^{l, {n_{\text{h}}}}\}_{ {n_{\text{h}}}=1}^{ {N_{\text{h}}}}, F^{l-1})$ is also assumed 
as in \citet{chen2023calibrating}.
Moreover, we also follow the assumption in \citet{chen2023calibrating} on 
the conditional independency for each head across layers.
When considering $\{\bm{u}^{l, {n_{\text{h}}}}\}_{ {n_{\text{h}}}=1}^{ {N_{\text{h}}}}$ from each head, we then have the factorizations:
\begin{align} \label{eq::heads}
    \mathrm{p}\left(\{\bm{u}^{l, {n_{\text{h}}}}\}_{ {n_{\text{h}}}=1}^{ {N_{\text{h}}}})|F^{l-1}\right)
    =\prod\nolimits_{ {n_{\text{h}}}=1}^{ {N_{\text{h}}}}\mathrm{p}\left(\bm{u}^{l, {n_{\text{h}}}}|F^{l-1}\right),
    \quad
    \mathrm{q}\left(\{\bm{u}^{l, {n_{\text{h}}}}\}_{ {n_{\text{h}}}=1}^{ {N_{\text{h}}}})|F^{l-1}\right)
    =\prod\nolimits_{ {n_{\text{h}}}=1}^{ {N_{\text{h}}}}\mathrm{q}\left(\bm{u}^{l, {n_{\text{h}}}}|F^{l-1}\right).
\end{align}

With the prerequisites derived above, we now proceed to formulate the ELBO in our KEP-SVGP:
\begin{align}\label{eq:our:elbo:derivation}
    & \mathcal{L}_{\text{ELBO}}
     \overset{\eqref{eq::dev_elbo}}{=}
    \mathbb{E}_{\mathrm{q}\left(\{F^l\}_{l=1}^L, \{\bm{u}^{l, {n_{\text{h}}}}\}_{l=1, {n_{\text{h}}}=1}^{L, {N_{\text{h}}}}|F^0\right)}
    \left[\log\frac{\mathrm{p}\left(Y,\{F^l\}_{l=1}^L, \{\bm{u}^{l, {n_{\text{h}}}}\}_{l=1, {n_{\text{h}}}=1}^{L, {N_{\text{h}}}}|F^0\right)}{\mathrm{q}\left(\{F^l\}_{l=1}^L, \{\bm{u}^{l, {n_{\text{h}}}}\}_{l=1, {n_{\text{h}}}=1}^{L, {N_{\text{h}}}}|F^0\right)}
    \right]
    \vspace{0.15cm}
    \nonumber\\
    & \overset{\eqref{eq::joint_density},\eqref{eq::var_post}}{=}  \mathbb{E}_{\mathrm{q}\left(\{F^l\}_{l=1}^L, \{\bm{u}^{l, {n_{\text{h}}}}\}_{l=1, {n_{\text{h}}}=1}^{L, {N_{\text{h}}}}|F^0\right)} 
    \left[\log
\frac{\mathrm{p}\left(Y|F^L\right)\prod\nolimits_{l=1}^L\mathrm{p}\left(F^l| \{\bm{u}^{l, {n_{\text{h}}}}\}_{ {n_{\text{h}}}=1}^{ {N_{\text{h}}}},F^{l-1}\right)\mathrm{p}\left(\{\bm{u}^{l, {n_{\text{h}}}}\}_{ {n_{\text{h}}}=1}^{ {N_{\text{h}}}})|F^{l-1}\right)}{\prod\nolimits_{l=1}^{L}\mathrm{p}(F^l|\{\bm{u}^{l, {n_{\text{h}}}}\}_{ {n_{\text{h}}}=1}^{ {N_{\text{h}}}}, F^{l-1})
    \mathrm{q}\left(\{\bm{u}^{l, {n_{\text{h}}}}\}_{ {n_{\text{h}}}=1}^{ {N_{\text{h}}}})|F^{l-1}\right)}
    \right]
    \vspace{0.15cm}
    \nonumber\\
    & \overset{\eqref{eq::heads}}{=} \mathbb{E}_{\mathrm{q}\left(\{F^l\}_{l=1}^L, \{\bm{u}^{l, {n_{\text{h}}}}\}_{l=1, {n_{\text{h}}}=1}^{L, {N_{\text{h}}}}|F^0\right)} 
    \left[\log\mathrm{p}\left(Y|F^L\right)\right]
    {+} \mathbb{E}_{\mathrm{q}\left(\{F^l\}_{l=1}^L, \{\bm{u}^{l, {n_{\text{h}}}}\}_{l=1, {n_{\text{h}}}=1}^{L, {N_{\text{h}}}}|F^0\right)} 
    \left[\log\frac{\prod\nolimits_{l=1, {n_{\text{h}}}=1}^{L, {N_{\text{h}}}}\mathrm{p}\left(\bm{u}^{l, {n_{\text{h}}}}|F^{l-1}\right)}{\prod\nolimits_{l=1, {n_{\text{h}}}=1}^{L, {N_{\text{h}}}}\mathrm{q}\left(\bm{u}^{l, {n_{\text{h}}}}|F^{l-1}\right)}\right]
    \vspace{0.15cm}
    \nonumber\\
    & = \mathbb{E}_{\mathrm{q}\left(F^L|F^0\right)} 
    \left[\log\mathrm{p}\left(Y|F^L\right)\right]
{+} \sum\nolimits_{l=1}^{L}\sum\nolimits_{ {n_{\text{h}}}=1}^{ {N_{\text{h}}}}
    \mathbb{E}_{\mathrm{q}\left(F^{l-1}\right)}
    \mathbb{E}_{\mathrm{q}\left(\bm{u}^{l, {n_{\text{h}}}}|F^{l-1}\right)} 
    \left[\log\frac{\mathrm{p}\left(\bm{u}^{l, {n_{\text{h}}}}|F^{l-1}\right)}{\mathrm{q}\left(\bm{u}^{l, {n_{\text{h}}}}|F^{l-1}\right)}\right]
    \vspace{0.15cm}
    \nonumber\\
    & = \mathbb{E}_{\mathrm{q}\left(F^L|F^0\right)} 
    \left[\log\mathrm{p}\left(Y|F^L\right)\right]
    - \sum\nolimits_{l=1}^{L}\sum\nolimits_{ {n_{\text{h}}}=1}^{ {N_{\text{h}}}}
    \mathbb{E}_{\mathrm{q}\left(F^{l-1}\right)}
    \left[\text{KL}\left(\mathrm{q}\left(\bm{u}^{l, {n_{\text{h}}}}|F^{l-1}\right) \|\,\mathrm{p}\left(\bm{u}^{l, {n_{\text{h}}}}|F^{l-1}\right) \right)\right],
\end{align}
where
$\mathrm{q}(F^L|F^0)=\int \prod_{l=1}^L
\mathrm{p}(F^l|\{\bm{u}^{l, {n_{\text{h}}}}\}_{ {n_{\text{h}}}=1}^{ {N_{\text{h}}}},F^{l-1})
\mathrm{q}(\{\bm{u}^{l, {n_{\text{h}}}}\}_{ {n_{\text{h}}}=1}^{ {N_{\text{h}}}} | F^{l-1})\,
\{\mathrm{d}\bm{u}^{l, {n_{\text{h}}}}\}_{l=1, {n_{\text{h}}}=1}^{L, {N_{\text{h}}}}
\{\mathrm{d}F^{l}\}_{l=1}^{L-1}$.
In this regard, the first term in the above ELBO can be estimated using Monte-Carlo samples layer-wise with the
reparameterization trick \cite{kingma2013auto}.
As introduced in Section \ref{sec:svgp:eigen:features}, we consider the independent multi-output Gaussian Processes \cite{leibfried2020tutorial} by specifying 
separate $s$ single-output SVGPs. 
Hence, an independent SVGP is formulated for each of the output dimension, i.e., $\mathrm{q}(\bm{u}^{l, {n_{\text{h}}}}_{[d]}|F^{l-1})=\mathcal{N}(\bm{m}_{\bm{u},[d]},S_{\bm{uu},[d]})$, 
$\mathrm{p}(\bm{u}^{l, {n_{\text{h}}}}_{[d]}|F^{l-1})=\mathcal{N}(0,\Lambda^2)$. Thus,  the KL-divergence term in \eqref{eq:our:elbo:derivation} for each head index by $ {n_{\text{h}}}$ in each layer indexed by $l$ can be expressed in the following form:
\begin{align*}
     &\text{KL}\left(\mathrm{q}\left(\bm{u}^{l, {n_{\text{h}}}}|F^{l-1}\right) \|\,\mathrm{p}\left(\bm{u}^{l, {n_{\text{h}}}}|F^{l-1}\right) \right)
    = \sum\nolimits_{d=1}^s 
    \text{KL}\left(\mathrm{q}\left(\bm{u}^{l, {n_{\text{h}}}}_{[d]}|F^{l-1}\right) \|\,\mathrm{p}\left(\bm{u}^{l, {n_{\text{h}}}}_{[d]}|F^{l-1}\right) \right)
    \\
    & 
    \overset{\eqref{eq::e:r:score_sgp},\eqref{eq::same_var_dis}}{=} \frac{1}{2}\sum\nolimits_{d=1}^s
    \left[\text{Tr}(\Lambda^{-2}S_{\bm{uu},[d]}) + \bm{m}_{\bm{u},[d]}^\top\Lambda^{-2}\bm{m}_{\bm{u},[d]}+\log\frac{|\Lambda^{2}|}{|S_{\bm{uu},[d]}|} - s\right] {,}
\end{align*}
where $s$ is the number of output dimensions of $\bm{u}^{l, {n_{\text{h}}}}_{[d]}\in\mathbb{R}^s$, and also the number of ``inducing points'' for the SVGPs.  
This completes the derivation of \eqref{eq::elbo_spg}.

\section{More Background Materials}
\label{sec::background::appendix}
In this section, we recall \citet{suykens2016svd,chen2023primal} so as to provide a better understanding of KEP-SVGP.

\paragraph{SVD under LS-SVM framework \cite{suykens2016svd}}
Let $X\in\mathbb{R}^{N\times M}$ be the data matrix, we first define two sources of data, corresponding to the rows and columns of the data matrix $X$ respectively: $\{\bm{x}_i:=X^\top\epsilon_i\}_{i=1}^N$, $\{\bm{z}_j:=X\varepsilon_j\}_{j=1}^M$ where $\epsilon_i\in\mathbb{R}^N$, $\varepsilon_j\in\mathbb{R}^M$ denote standard basis vectors, that is, the column vectors of the identity matrices $I_N$ and $I_M$, respectively.
The operations from $X$ to $\{\bm{x}_i\in\mathbb{R}^M\}_{i=1}^N$ and $\{\bm{z}_j\in\mathbb{R}^N\}_{j=1}^M$ can be denoted by \textbf{two linear feature maps}:
\begin{align}\label{eq:svd:two_feature_maps}
    \varphi(\bm{x}_i):=C^\top\bm{x}_i=C^\top X^\top\epsilon_i,
    \quad
    \psi(\bm{z}_j):= \bm{z}_j = X\varepsilon_j,
\end{align}
where $\varphi:\mathbb{R}^M\to\mathbb{N}$, $\psi:\mathbb{R}^N\to\mathbb{R}^N$ and $C\in\mathbb{R}^{M\times N}$ is a compatibility matrix so that $\bm{x}_i$, $\bm{z}_j$ can be compared with each other after applying he feature maps. 
Now we consider the following constrained optimization problem, i.e., primal problem, which aims at maximizing the projection variances w.r.t. rows and columns data of $X$:
\begin{equation}\label{eq:ksvd:original}
\begin{array}{rl}
    \max\limits_{w, v, e_i, r_j }&  J =  \dfrac{1}{2\gamma}\sum\nolimits_{i=1}^N e_i^2 + \dfrac{1}{2\gamma}\sum\nolimits_{j=1}^M r_j^2 - w^\top v  \\
    {\rm s.t.} &  
    {e_i = w^\top \varphi(\bm{x}_i), \  i=1, \ldots, N,}
    \\
    & 
    {r_j = v^\top \psi(\bm{z}_j), \ j=1, \ldots, M,}
\end{array}
\end{equation}
where $w,v\in\mathbb{R}^{N}$ are the projection weights, $e_i, r_j\in\mathbb{R}$ are the projection scores, and $\gamma\in\mathbb{R}$ is the regularization coefficient.
This projection variances maximization in \eqref{eq:ksvd:original} exactly follows the spirit of singular value decomposition (SVD).

\paragraph{KSVD problem for self-attention \cite{chen2023primal}}
Let $\{\bm{x}_i\in\mathbb{R}^d\}_{i=1}^N$ be the input data sequence. 
The asymmetric self-attention kernel matrix $K_{\text{att}}$ can be formulated by $K_{\text{att}}:=[\kappa_{\text{att}}(\bm{x}_i,\bm{x}_j)]\in\mathbb{R}^{N\times N}$ where $\kappa_{\text{att}}(\bm{x}_i,\bm{x}_j):=\langle\phi_q(\bm{x}_i),\phi_k(\bm{x}_j)\rangle$ with two feature maps $\phi_q, \phi_k$ related to queries and keys, corresponding to \eqref{eq:svd:two_feature_maps}. 
\citet{chen2023primal} gives the \textbf{primal problem} of KSVD of the self-attention mechanism as follows:
\begin{equation}\label{eq:ksvd:primal_attention}
\begin{array}{rl}
    \max\limits_{W_e, W_r, \Lambda }&  J =  \dfrac{1}{2}\sum\nolimits_{i=1}^N e(\bm{x}_i)^\top\Lambda^{-1} e(\bm{x}_i) + \dfrac{1} {2}\sum\nolimits_{j=1}^N  r(\bm{x}_j)^\top \Lambda^{-1} r(\bm{x}_j) - \text{Tr}\left(W_e^\top W_r\right)  \\
    {\rm s.t.} &  
    {e(\bm{x}_i) = W_e^\top \phi_q(\bm x_i), \  i=1, \ldots, N,}
    \\
    & 
    {r(\bm{x}_j) = W_r^\top \phi_k(\bm x_j), \ j=1, \ldots, N,}
\end{array}
\end{equation}
which corresponds to the primal problem in \eqref{eq:ksvd:original}.
Here, $W_e,W_r \in \mathbb{R}^{p\times s}$ are the projection weights,
$\phi_q(\cdot), \phi_k(\cdot)\colon \mathbb{R}^d \to \mathbb{R}^p$ are the feature maps, 
$e(\bm{x}_i)=W_e^\top \phi_q(\bm x_i)\in \mathbb{R}^{s}$, $r(\bm{x}_j)=W_r^\top \phi_k(\bm x_j) \in \mathbb{R}^{s}$ are the projection scores,
and $\Lambda \in \mathbb{R}^{s\times s}$ is the regularization coefficient 
which is a positive diagonal matrix. 
$J$ in \eqref{eq:ksvd:primal_attention}
maximizes the projection variances of $W_e^\top\phi_q(\bm{x}_i)$, $W_r^\top\phi_k(\bm{x}_j)$ regarding queries and keys, and involves a regularization term coupling the projections.

With Lagrangian duality and KKT conditions, the \textbf{dual problem} of \eqref{eq:ksvd:primal_attention} is
\begin{align} \label{eq::dual_problem::primal_attention}
    K_{\text{att}}H_r=H_e\Lambda,\quad K_{\text{att}}^\top H_e=H_r\Lambda
\end{align}
where $\Lambda \in \mathbb R^{s\times s}$ is a positive diagonal matrix serves as singular values of $K_{\text{att}}$, and  $H_e = [\bm h_{e_1}, \ldots, \bm h_{e_N}]^\top \in \mathbb R^{N\times s}$, 
$H_r = [\bm h_{r_1}, \ldots, \bm h_{r_{N}}]^\top \in \mathbb R^{{N}\times s}$ are the dual variables serving as the left and right singular vectors, respectively.  
Then the projection scores can be either represented in the primal using explicit feature maps or in the dual using kernel functions:
\begin{equation*}
    \begin{array}{rll}
         \text{Primal:} & 
         \left\{
         \begin{array}{l}
            {e}(\bm{x}) = W_e^\top  \phi_q(\bm{x})
            \\
            {r}(\bm{x}) = W_r^\top   \phi_k(\bm{x})
         \end{array},
         \right.
        \quad
         \text{Dual:} &
         \left\{
         \begin{array}{l}
            {e}(\bm{x})  = \sum\nolimits_{j=1}^N   \bm  h_{r_j}    \kappa_{\text{att}}(\bm{x},\bm{x}_j) 
            \\
            {r}(\bm{x}) = \sum\nolimits_{i=1}^N  \bm   h_{e_i}    \kappa_{\text{att}}(\bm{x}_i,\bm{x}).
         \end{array}
         \right.
    \end{array}
\end{equation*}
According to Lemma 4.2 in \citet{chen2023primal}, the solutions $H_e, H_r, \Lambda$ to the dual problem \eqref{eq::dual_problem::primal_attention} lead to the zero-value objective $J$ in \eqref{eq:ksvd:primal_attention}.
Therefore, we set $\mathcal{L}_{\text{KSVD}}:=J^2$ in \eqref{eq::ksvd_loss} as our KSVD regularization loss.
By minimizing $\mathcal{L}_{\text{KSVD}}$ to zero through SGD-based optimizers, we realize SVD on $K_{\text{att}}$ in self-attention in an approximate way.

\section{More Experiment Details}
First of all, in addition to reporting good performance on in-distribution datasets, we would like to provide some rationales behind KEP-SVGP's good performance in distribution-shift robustness and out-of-distribution detection, which are two common tasks for Bayesian models:
\begin{itemize}
    \item With KSVD, we use the pair of adjoint eigenfunctions of the attention kernel as the ``inducing features'' in two SVGPs with two benefits: \textit{i)} eigenfunctions span an orthogonal system seeking informative feature spaces; \textit{ii)} eigenfunctions in our KSVD promote low-rank property to attention, where noisy patterns could be filtered out.
    \item Recall that the distribution-shift data corrupts the clean data with shifts, while out-of-distribution (OOD) data are from another different distribution. \textit{i)} As distribution shifts can serve as noisy patterns, the low-rank property can help filtering out these noises. \textit{ii)} As the feature space of OOD data can be largely different from that of the in-distribution data, the informative features by the eigenfunctions can help distinguish such differences.
\end{itemize}

\subsection{More Details on Experimental Setups}\label{sec:setups:appendix}
All experiments presented in this work are implemented with PyTorch, which can be conducted on a single NVIDIA GeForce RTX 2070 SUPER GPU.
Our implementation is available at \href{https://github.com/yingyichen-cyy/KEP-SVGP}{https://github.com/yingyichen-cyy/KEP-SVGP}.

\paragraph{Experiments on CIFAR-10 and CIFAR-100}
For both CIFAR-10 and CIFAR-100, we randomly split the original training set into 90\% training and 10\% validation set, leading to a training set of 45K samples and a validation set of 5K.
The test set is of 10K samples.
For both datasets, we use 
7-layer ViT \cite{dosovitskiy2021an} where the 32$\times$32 input images are tokenized with patches of size $4\times4$, the embedding dimension is 384, the hidden dimension is 384, {the number of heads} is 12, the dropout ratio is 0.1, and the classification token is turned off.
For all KEP-SVGPs on these two datasets, we set the regularization constant of KSVD loss in our objective $\min \,-\mathcal{L}_{\text{ELBO}} + \eta \mathcal{L}_{\rm KSVD}$ as $\eta=10$, set the rank for KSVD as $s=10$.
In our experimental section, we choose the feature maps related to the cosine similarity kernel on queries and keys as in \citet{chen2023primal}.
All models are trained from scratch with ADAM optimizer \cite{kingma2014adam}, except the post-hoc methods including Temperature Scaling and KEFLLLA, for 300 epochs with 5 warm-up epochs. The batch size is 128, and a cosine learning rate schedule is utilized with a learning rate of $10^{-3}$ and minimum learning rate of $10^{-5}$.
Ensemble methods are based on the models trained independently over 5 trials.
More specifically, deep Ensembles is implemented by the mean ensembling of 5 independently trained transformers, which is the referred ``regular transformer-based model along with ensembling techniques''. 
KEP-SVGP Ensembles does the same for 5 independently trained KEP-SVGP transformers.
The best models are selected with the best validation accuracy.
During inference, for MC Dropout and our KEP-SVGP, predictive uncertainty is estimated using 10 Monte Carlo samples.
Note that SGPA \cite{chen2023calibrating} is very time and memory consuming with 7-layer architectures, therefore we do not include it in the comparisons on CIFAR datasets in Table \ref{tab::in_dist}.
However, we do include SGPA in Table \ref{tab::efficiency} with all models trained with the same architectures as done in its original paper \cite{chen2023calibrating} for fair comparisons: 5-layer ViT on CIFAR-10, 6-layer ViT on CIFAR-100.

\paragraph{Experiments on IMDB}
We randomly split the IMDB original training set into 35K as training and 5K as validation, the test set is of 10K samples.
IMDB is with the maximum sequence length of 512.
Following \citet{chen2023calibrating}, we use 1-layer Transformer \cite{vaswani2017attention} where the embedding dimension is 128, the hidden dimension is 128, the number of heads is 8, and the dropout ratio is 0.1.
For our KEP-SVGP with addition scheme \eqref{eq:merge:output} for merging of SVGPs outputs, we set KSVD regularization constant as $\eta=10$, the KSVD rank as $s=10$, with feature maps related to the cosine similarity kernel on queries and keys \cite{chen2023primal}.
We train all models with ADAM optimizer, except for the post-hoc methods including Temperature Scaling and KFLLLA, for 20 epochs with 5 warm-up epochs, a batch size of 32, and a initial learning rate $10^{-3}$ which decays to $10^{-4}$ following a cosine learning rate decay.
Ensemble methods are based on the models trained independently over 5 trials.
The best models are selected with the best validation accuracy.
During inference, for MC Dropout, SGPA and our KEP-SVGP, predictive uncertainty is estimated with 10 Monte Carlo samples.

\paragraph{Experiments on CoLA}
This dataset provides an in-distribution training with 8551 samples and a in-distribution test of 527.
Following \citet{chen2023calibrating}, we use 2-layer Transformer \cite{vaswani2017attention} where the embedding dimension is 128, the hidden dimension is 256, the number of heads is 4.
For the input embedding, we adopt ELMO-style representation \cite{peters2018deep}.
For our KEP-SVGP with addition merging scheme in \eqref{eq:merge:output}, we set KSVD regularization constant as $\eta=1$, the KSVD rank as $s=5$, 
with feature maps $\phi_q$, $\phi_k$ in \eqref{eq::primal_dual} related to the cosine similarity kernel \cite{chen2023primal}.
We train all models with ADAM optimizer, except for the post-hoc methods including Temperature Scaling and KFLLLA, for 50 epochs with 5 warm-up epochs, a batch size of 32, and a initial learning rate $5\times10^{-4}$ which decays to $10^{-5}$ following a cosine learning rate decay.
Ensemble methods are based on the models trained independently over 5 trials.
During inference, for MC Dropout, SGPA and our KEP-SVGP, predictive uncertainty is estimated using 10 Monte Carlo samples.

\subsection{More Details on Concatenation Weights}
\label{sec:ab:appendix}
For the concatenation merging scheme in Section \ref{sec:svgp:eigen:features}, we can replace $W_1^{\text{cat}}\in\mathbb{R}^{N\times 2N}$ with $AB^\top$ where $A\in\mathbb{R}^{N\times s}$, $B\in\mathbb{R}^{2N\times s}$ to maintain the overall linear time complexity with $N$ when computing KEP-SVGP.
We provide empirical evaluations on comparing these two concatenation weight schemes in Table \ref{tab::diff_concate_weight}. 
By utilizing $AB^\top$, comparable performances are obtained at less computational cost. 
However, we still adopt $W_1^\text{cat}$ for all experiments in this paper so as to maintain good failure prediction and calibration performances.
\begin{table}[t]
    \caption{Comparisons on $W_1^{\text{cat}}\in\mathbb{R}^{N\times 2N}$ and $AB^\top$, $A\in\mathbb{R}^{N\times s}$, $B\in\mathbb{R}^{2N\times s}$ on CIFAR-10. Forward time (s) is on a single V100.}
    \label{tab::diff_concate_weight}
    \vspace{-2mm}
    \begin{center}
    \resizebox{\textwidth}{!}{
    \begin{tabular}{ccccccccc}
    \toprule
    Concatenation & Forward Time (s) & ACC $\uparrow$ & AURC $\downarrow$ & AUROC $\uparrow$ & FPR95 $\downarrow$ & ECE $\downarrow$ & NLL $\downarrow$ & Brier $\downarrow$
    \\ \midrule
    $W_1^{\text{cat}}\in\mathbb{R}^{N\times 2N}$ 
    & 0.08$\pm$0.03 & 84.70$\pm$0.61 & \textbf{35.15}$\pm$2.65 & \textbf{87.20}$\pm$0.65 & \textbf{64.93}$\pm$1.41 & 10.60$\pm$0.45 & \textbf{8.00}$\pm$0.56 & \textbf{25.45}$\pm$1.05 
    \\
    $AB^\top$, $A\in\mathbb{R}^{N\times s}$, $B\in\mathbb{R}^{2N\times s}$ 
    & \textbf{0.05}$\pm$0.02 
    & \textbf{84.71}$\pm$0.10 & 36.01$\pm$1.01 & 86.63$\pm$0.31 & 67.72$\pm$1.08 & \textbf{10.53}$\pm$0.22 & 8.03$\pm$0.32 & 25.57$\pm$0.33
    \\ \bottomrule
    \end{tabular}}
    \end{center}
    \vskip -0.2in
\end{table}

\subsection{More Details on Calibration Improvement}
\label{sec:cal_boost:appendix}
We discuss the further potentials of KEP-SVGP here.
The benefits of our KEP-SVGP can beyond the existing evaluations, as the key idea of fully utilizing the asymmetry in self-attention is compatible with other calibration method and thus serves as a good complement for a further boost in performances, as demonstrated in Table \ref{tab::cali_improve_appendix}.
Specifically, we find that
\begin{itemize}
    \item KEP-SVGP$+$TS on IMDB: Temperature Scaling (TS) is a post-hoc method tuning the temperature scale inside the softmax probability in the classification head. As KEP-SVGP directly models the attention blocks, it can be combined with TS, leading to an improved calibration.
    \item KEP-SVGP$+$KFLLLA on CoLA: KFLLLA is a post-hoc method working on the last linear layer in the classification head. Hence, fine-tuning the classification head with KFLLLA on pre-trained KEP-SVGP transformer improves both failure prediction and calibration metrics.
\end{itemize}
\begin{table*}[t]
    \caption{KEP-SVGP serves as a good complement to other methods for improving calibration.
    Experimental results are reported over five trials, with the best mean results shown in bold.
    ACC, AUROC, FPR95, ECE and Brier are percentages, 
    AURC is $\times 10^3$, NLL is $\times 10$.
    }
    \vspace{-3mm}
    \label{tab::cali_improve_appendix}
    \begin{center}
    \resizebox{\textwidth}{!}{
    \begin{tabular}{ccccccccc}
        \toprule
        Dataset & Method & ACC/MCC $\uparrow$ & AURC $\downarrow$ & AUROC $\uparrow$ & FPR95 $\downarrow$ & ECE $\downarrow$ & NLL $\downarrow$ & Brier $\downarrow$ 
        \\ \midrule
        \multirow{4}{*}{\begin{tabular}[c]{@{}c@{}}IMDB\\ \cite{maas2011learning}\end{tabular}} 
        & MSP \cite{hendrycks2016baseline}             
        & 88.17$\pm$0.52 & 35.27$\pm$3.04 & 82.29$\pm$0.87 & 71.41$\pm$1.57 & 4.01$\pm$1.36 & 3.10$\pm$0.26 & 17.88$\pm$0.95         
        \\
        & Temperature Scaling (TS) \cite{guo2017calibration}              
        & 88.17$\pm$0.52 & 35.27$\pm$3.04 & 82.29$\pm$0.87 & 71.08$\pm$1.55 & {\bf 1.05}$\pm$0.70 & 2.89$\pm$0.12 & 17.40$\pm$0.80 
        \\ \cmidrule(lr){3-9} 
        & KEP-SVGP (ours)          
        & 89.01$\pm$0.14 & {\bf 30.69}$\pm$0.69 & {\bf 83.22}$\pm$0.31 & {\bf 68.15}$\pm$0.95 & 3.72$\pm$0.81 & 3.00$\pm$0.13 & 16.56$\pm$0.25  
        \\
        & KEP-SVGP$+$TS (ours)   
        & {\bf 89.01}$\pm$0.14 & 30.71$\pm$0.76 & {\bf 83.22}$\pm$0.36 & 68.38$\pm$0.60 & 1.19$\pm$0.19 & {\bf 2.73}$\pm$0.03 & {\bf 16.21}$\pm$0.16
        \\ \midrule
        \multirow{4}{*}{\begin{tabular}[c]{@{}c@{}}CoLA\\ \cite{warstadt2019neural}\end{tabular}} 
        & MSP \cite{hendrycks2016baseline}             
        & 26.93$\pm$1.38 & 205.47$\pm$7.62 & 64.55$\pm$0.86 & 89.86$\pm$1.29 & 23.84$\pm$2.23 & 14.45$\pm$2.83 & 52.15$\pm$2.43 
        \\
        & KFLLLA \cite{kristiadi2020being}
        & 26.90$\pm$1.31 & 204.31$\pm$8.57 & 64.60$\pm$0.96 & 90.06$\pm$0.74 & {\bf 2.51}$\pm$1.09 & 5.94$\pm$0.04 &  40.52$\pm$0.38
        \\ \cmidrule(lr){3-9} 
        & KEP-SVGP (ours)          
        & 30.54$\pm$1.61 & 186.66$\pm$8.50 & 65.16$\pm$0.86 & 88.39$\pm$0.83 & 15.89$\pm$3.48 & 8.54$\pm$1.66 & 43.55$\pm$2.99 
        \\ 
        & KEP-SVGP$+$KFLLLA (ours)  
        & {\bf 31.22$\pm$1.63} & {\bf 185.58}$\pm$6.57 & {\bf 65.49}$\pm$1.31 & {\bf 87.97}$\pm$2.12 & 5.81$\pm$2.19 & {\bf 5.78}$\pm$0.04 & {\bf 38.98}$\pm$0.44
        \\ \bottomrule
    \end{tabular}}
    \end{center}
    \vspace{-5mm}
\end{table*}

\subsection{More Details on Time Efficiency}
\label{sec:time_eff:appendix}
In addition to Table \ref{tab::efficiency}, we also report forward time in seconds during training in Table \ref{tab::efficiency_appendix} so as not to include the time taken by the optimizer.
It can be seen that we reach the same conclusion as in Section \ref{subsec::time_efficiency} that KEP-SVGP distinctively outperforms MSP on all datasets with only a slightly extra forward time during training.

\begin{table}[t]
    \caption{Performance and forward time (s) on a single NVIDIA Tesla V100 SXM2 32 GB. Results are reported over five trials.}
    \label{tab::efficiency_appendix}
    \vspace{-2mm}
    \begin{center}
    \resizebox{\textwidth}{!}{
    \begin{tabular}{cccccccccc}
        \toprule
        \multirow{2}{*}{Method} 
        & \multirow{2}{*}{\begin{tabular}[c]{@{}c@{}}Time\\Complexity\end{tabular}} 
        & \multicolumn{2}{c}{CIFAR-10}
        & \multicolumn{2}{c}{CIFAR-100}
        & \multicolumn{2}{c}{IMDB}  
        & \multicolumn{2}{c}{CoLA}
        \\ \cmidrule(lr){3-4} \cmidrule(lr){5-6} \cmidrule(lr){7-8} \cmidrule(lr){9-10}
        & & ACC $\uparrow$ & Forward Time (s)
        & ACC $\uparrow$ & Forward Time (s)
        & ACC $\uparrow$ & Forward Time (s)
        & MCC $\uparrow$ & Forward Time (s)
        \\ \midrule
        MSP \cite{hendrycks2016baseline} 
        & $\mathcal{O}(N^2)$
        & 78.11$\pm$0.10 & {\bf 0.02}$\pm$0.01 
        & 52.16$\pm$0.50 & {\bf 0.02}$\pm$0.02
        & 88.17$\pm$0.52 & {\bf 0.001}$\pm$0.0
        & 26.93$\pm$1.38 & 0.40$\pm$0.33        
        \\
        SGPA \cite{chen2023calibrating}  
        & $\mathcal{O}(N^2s)$
        & 77.87$\pm$0.12 & 0.15$\pm$0.0
        & 53.02$\pm$0.36 & 0.27$\pm$0.0
        & 88.36$\pm$0.75 & 0.63$\pm$0.07
        & 26.15$\pm$1.12 & 0.47$\pm$0.37      
        \\ \cmidrule(lr){2-10}
        KEP-SVGP (ours) 
        & {$\mathcal{O}(Ns^2)$}
        & {\bf 78.27}$\pm$0.30 & {\bf 0.02}$\pm$0.02
        & {\bf 56.26}$\pm$0.70 & 0.03$\pm$0.01
        & {\bf 89.01}$\pm$0.14 & 0.01$\pm$0.0
        & {\bf 30.54}$\pm$1.61 & {\bf 0.37}$\pm$0.41
        \\ \bottomrule   
    \end{tabular}}
    \end{center}
    \vskip -0.2in
\end{table}

\subsection{Additional Ablations}\label{appdx::ablations}
There are two main hyper-parameters in KEP-SVGP: the  {regularization constant} of KSVD loss $\eta$, the rank of KSVD $s$.
For the rank $s$, we adopt the default settings in \citet{chen2023primal} and set $s\in\{5, 10\}$.
As $\eta$ balances $\mathcal{L}_{\text{ELBO}}$ and $\mathcal{L}_{\text{KSVD}}$, it is of importance for KEP-SVGP.
Therefore, we provide the ablation of $\eta$ with rank $s=10$ fixed on CIFAR-10 \cite{krizhevsky2009learning}, 
and also the ablation on the addition and concatenation merging schemes given in \eqref{eq:merge:output} in Table \ref{tab::ablation_eta}, so as to give a guideline of the choice of schemes.

It can be seen in Table \ref{tab::ablation_eta} that the concatenation merging scheme has an overall better performance among all metrics than the addition scheme.
The possible reason can be that positive and negative output values from the two SVGPs branches can cancel each other out when employing the addition scheme, while the concatenation scheme can 
preserve the outputs information.
However, when dealing with language modelling tasks, we employ the addition scheme since it is sequence length independent while most of the language datasets are with varying sequence length.
We also find that $\eta=10$ returns better performances than choosing a small one.
A larger $\eta$ can help the model to conduct effective KSVD in an early stage, contributing to the construction of more accurate SVGPs branches and thereby leading to better overall performances.
\begin{table}[t]
    \caption{Ablation on KSVD regularization constant $\eta$ and the merging scheme of the SVGP pair on CIFAR-10.}
    \label{tab::ablation_eta}
    \vspace{2mm}
    \begin{center}
    \resizebox{0.57\columnwidth}{!}{
    \begin{tabular}{cccccccc}
        \toprule
        $\eta$ 
        & ACC $\uparrow$ 
        & AURC $\downarrow$ 
        & AUROC $\uparrow$& FPR95 $\downarrow$ & ECE $\downarrow$ & NLL $\downarrow$ & Brier $\downarrow$ 
        \\ \midrule
        \multicolumn{8}{c}{$[F^e_{[d]} ; F^r_{[d]}]$}      
        \\
        10 & \bf 84.70 & \bf 35.43 & 87.17 & \bf 62.55 & \bf 10.71 & \bf 7.95 & \bf 25.32   
        \\
        1 & 83.77 & 39.28 & 87.08 & 66.36 & 12.09 & 9.67 & 27.64    
        \\
        0.1 & 83.78 & 37.39 & \bf 87.86 & 64.12 & 12.04 & 9.62 & 27.46  
        \\
        0.01 & 82.60 & 44.60 & 86.17 & 67.13 & 13.12 & 11.16 & 29.85  
        \\ \midrule
        \multicolumn{8}{c}{$F^e_{[d]} + F^r_{[d]}$}                 
        \\
        10 & 84.04 & 38.99 & 86.64 & 66.04 & 11.64 & 9.39 & 27.02     
        \\
        1 & 84.45 & 37.32 & 86.97 & 65.02 & 11.30 & 8.80 & 26.27       
        \\
        0.1 & 83.56 & 39.74 & 87.03 & 63.63 & 12.22 & 9.49 & 27.84      
        \\
        0.01 & 81.46 & 49.71 & 85.81 & 66.51 & 13.94 & 11.24 & 31.35   
        \\ \bottomrule
    \end{tabular}}
    \end{center}
    \vspace{-5mm}
\end{table}

\subsection{Low-rank Property in Attention Kernel Matrix}\label{appdx::low_rank}
To further validate  {the approximate}
posterior distributions described 
in Remark \ref{rmk:approx}, we {present empirical evidence delving} into the low-rank property resided in the Transformer models.
{We adopt the two-layer Transformer on CoLA following the setups in Appendix \ref{sec:setups:appendix}.
Specifically, we consider 3 different models:
\textit{i)} MSP; 
\textit{ii)} last-layer KEP-SVGP;
\textit{iii)} two-layer KEP-SVGP.}
All our methods here are with addition merging scheme.
The spectrum analysis of the self-attention kernel matrix in each layer of each model is provided in the following Figure \ref{fig::low_rank}.
\begin{figure}[h]
    \begin{center}
    \centerline{\includegraphics[width=0.9\textwidth]{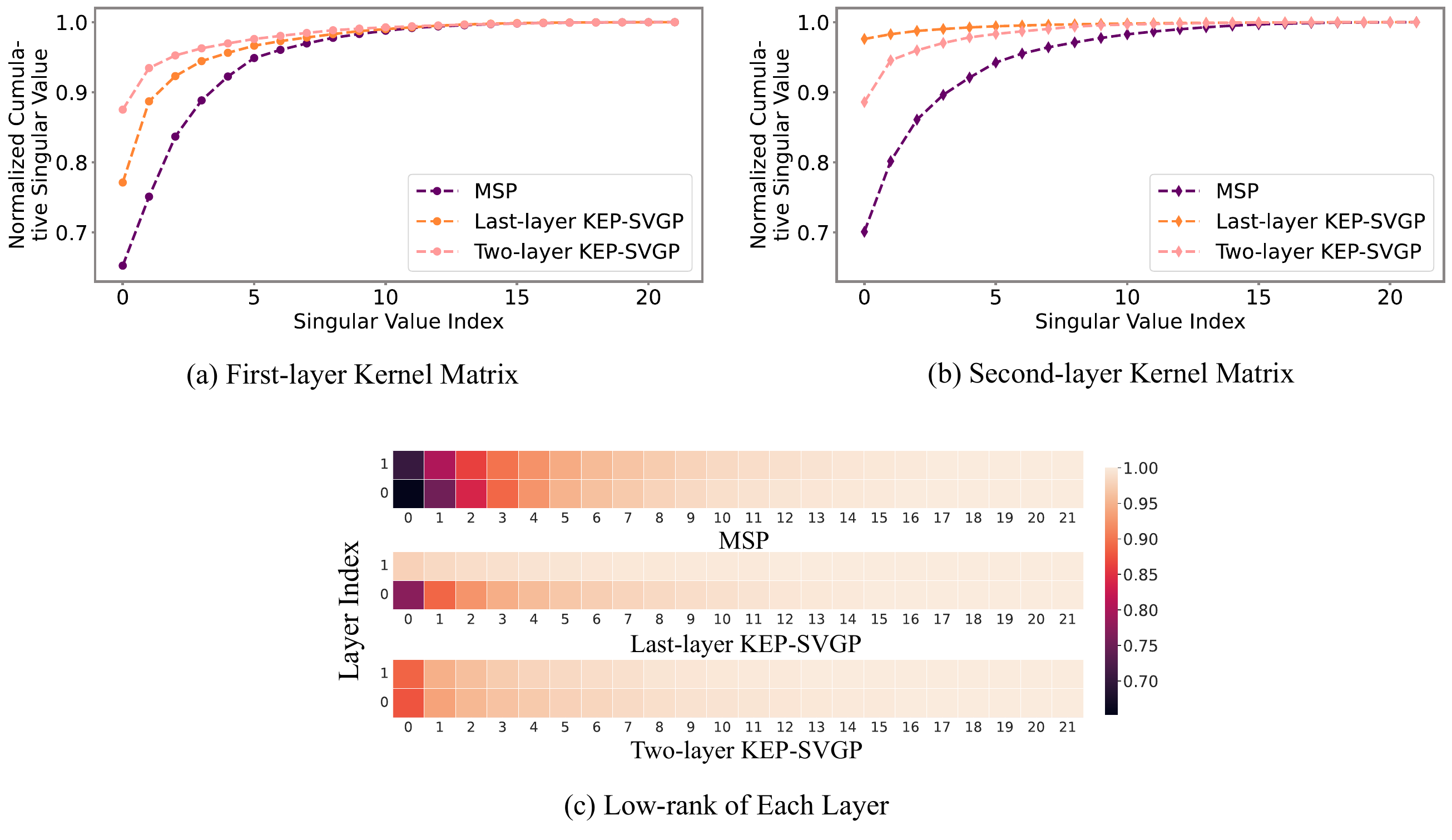}}
    \caption{{Spectrum analysis of the self-attention kernel matrix on CoLA. Specifically, we consider the normalized cumulative singular values of the attention matrix of the two-layer Transformer models.
    (a) plots the spectrum results of the first-layer kernel matrix;
    (b) plots the spectrum results of the second-layer kernel matrix;
    (c) plots the normalized cumulative singular values w.r.t. singular value index of each layer, showing the low-rank property of the attention matrix of each model.}}
    \label{fig::low_rank}
    \end{center}
    \vskip -0.2in
\end{figure}

According to the results in Figure \ref{fig::low_rank}, we find that
\vspace{-2mm}
\begin{itemize}
    \item 
    The attention matrix in the layers of Transformer has low-rank property, though
    the shallow layer may not 
    desires the low-rank property as much as the deeper layer. 
    This is consistent with the 
    findings in \citet{chen2023primal}. (Figure \ref{fig::low_rank}(c))
    \item Most of the information ($>95$\% explained variance)
    of the attention 
    matrix in both layers in the MSP baseline can be captured by the top-5 singular vectors. Thus, the hyperparameter $s=5$ in KSVD of our method is reasonable and approaches the ground-truth rank of the attention kernel.
    Our method captures distinctively higher explained variances  in the top singular vectors than the MSP baseline.
    (Figure \ref{fig::low_rank}(a), Figure \ref{fig::low_rank}(b))
    \item Applying KEP-SVGP
    only to the last layer
    also
    enhances the low-rank property of the attention  in other layers, as shown in the comparisons between our last-layer KEP-SVGP and two-layer KEP-SVGP.
    (Figure \ref{fig::low_rank}(a), Figure \ref{fig::low_rank}(b))
\end{itemize}

\subsection{Additional Visualization on Distribution-shift Data}
\begin{figure}[h]
    \begin{center}
    \centerline{\includegraphics[width=0.7\textwidth]{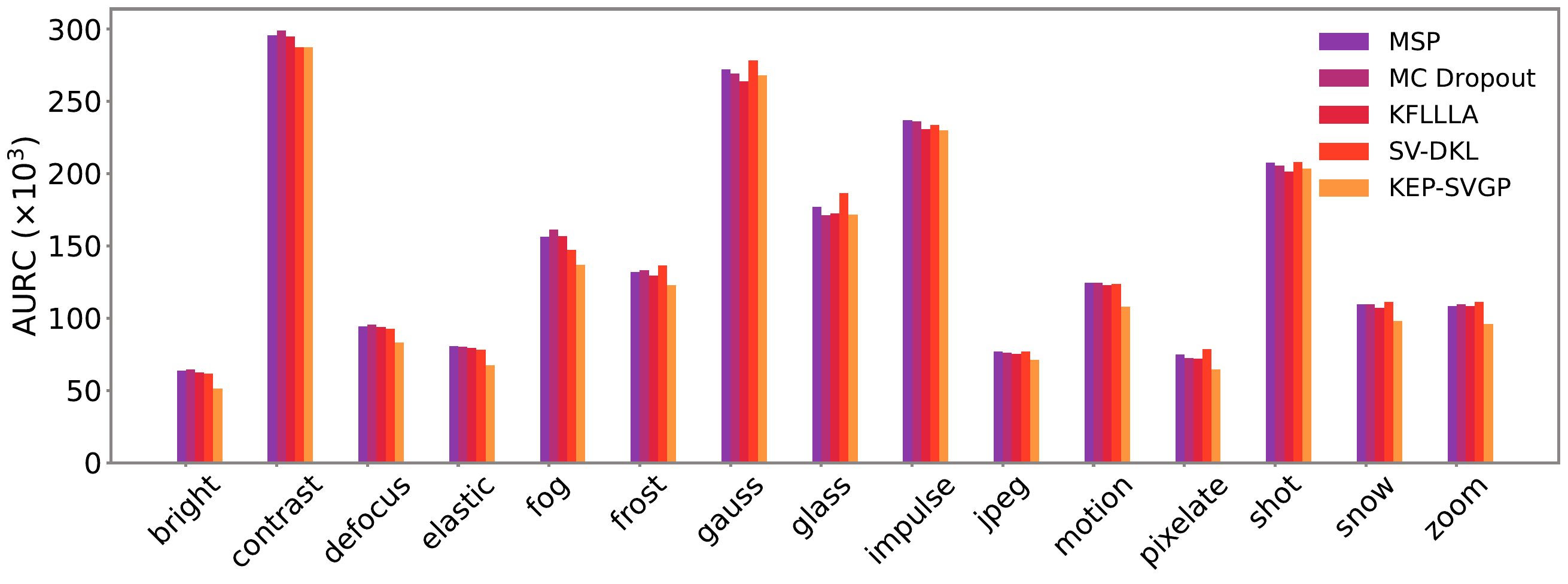}}
    \caption{Comparisons of our KEP-SVGP with baselines under distribution shift. Mean AURC results of all 5 severity levels under 15 types of corruption are reported, where models are trained on CIFAR-10/ViT and tested on CIFAR-10-C.}
    \label{fig::cifar10c_aurc}
    \end{center}
\end{figure}

We provide comparisons of our KEP-SVGP with all other baselines under distribution shift in Figure~\ref{fig::cifar10c_aurc}.
We report the mean AURC results of all 5 severity levels under 15 types of corruption on CIFAR-10-C~\cite{hendrycks2019benchmarking}.
All models are trained on CIFAR-10/ViT following the setups in Appendix \ref{sec:setups:appendix}, and then tested on CIFAR-10-C.
Our KEP-SVGP has overall better and stable AURC than all other comparing methods.

\end{document}